\documentclass[10pt,a4paper]{qwenreport}

\usepackage{docmute}

\reportlabel{Qwen Business Unit Technical Report}
\reportlogo{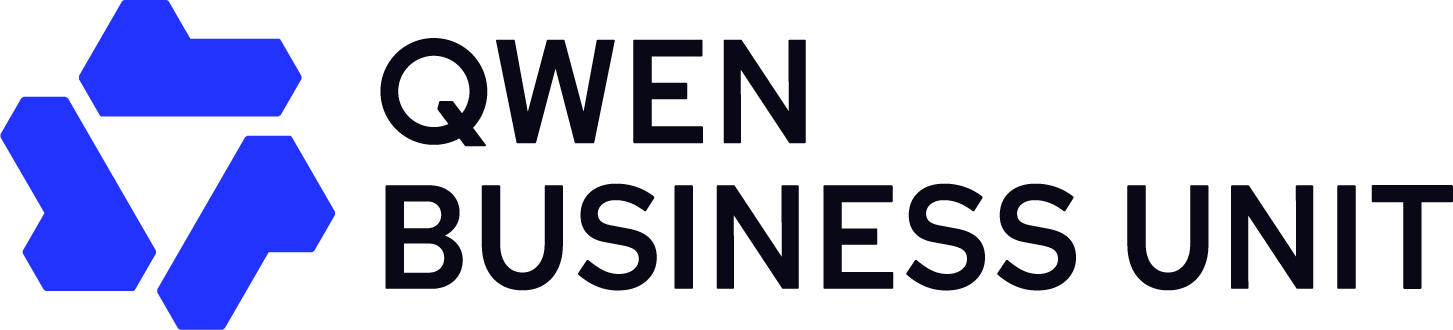}

\title{What to Edit Next: Visually Aligned Image-Editing Follow-Up Suggestions in Conversational Systems}

\author{%
  Zhijing Zhang\textsuperscript{1,2},\enspace
  Jinpeng Yu\textsuperscript{1,$\dagger$},\enspace
  Xin Song\textsuperscript{1},\enspace
  Bingnan Li\textsuperscript{1},\enspace
  Chuyue Li\textsuperscript{1,3},\enspace
  Changhui Du\textsuperscript{1},\enspace
  Yufeng Ai\textsuperscript{1},\enspace
  Xiaolin Fang\textsuperscript{2,*},\enspace
  Jiaming Liu\textsuperscript{1,*},\enspace
  Ruihua Huang\textsuperscript{1}\\
  \normalfont\fontsize{8}{10}\selectfont
  \textsuperscript{1}Qwen Business Unit of Alibaba,\enspace
  \textsuperscript{2}Southeast University,\enspace
  \textsuperscript{3}ShanghaiTech University
}

\newcommand{\reportfloatwidth}[1]{%
  \setlength{\columnwidth}{#1\textwidth}%
  \centering
}

\begin{document}

\begin{abstract}
Conversational assistants increasingly recommend follow-up edits to help users continue a task. Existing systems primarily target text-only interactions, leaving image-creation conversations underexplored. In image-creation tasks, useful follow-up edit suggestions must reflect user preferences, offer diverse directions, and remain executable on the current image. We collected 100,000 real multi-turn image-creation conversation samples from Qwen App and found that 80.1\% are image-dependent, underscoring the need for multimodal recommendation. We address this setting with a three-stage framework. In Stage 1, we use real online data to build a human-reviewed table of appropriate follow-up editing intents, then create SFT targets and fine-tune a multimodal policy. In Stage 2, to align rule-guided SFT suggestions with actual user choices, we use user click feedback to optimize the policy through multi-objective reinforcement learning. In Stage 3, to reduce visual inconsistencies between suggested edits and the current image, we introduce a visual verifier as additional training supervision. Extensive experiments demonstrate that our framework significantly outperforms baselines on both automatic and human evaluations. In a live user-randomized A/B test with millions of users, our final framework reduces visual inconsistency from 3.7\% to 0.9\%. Furthermore, it significantly improves recommendation CTR by 32.70\%, image take-away rate by 16.32\%, and average conversation turns per user by 39.90\% (all $p<0.05$). Project page: \hypersetup{urlcolor=blue}\url{https://what-to-edit-next.github.io/}\hypersetup{urlcolor=black}

\end{abstract}


\maketitle
\begingroup
\renewcommand{\thefootnote}{\fnsymbol{footnote}}
\footnotetext[1]{Co-corresponding authors. Emails: \texttt{jmliu1217@gmail.com}, \texttt{xiaolin@seu.edu.cn}.}
\footnotetext[2]{Project lead.}
\endgroup

\section{Introduction}
\label{sec:intro}

\begin{wrapfigure}{r}{0.46\textwidth}
\centering
\includegraphics[width=\linewidth]{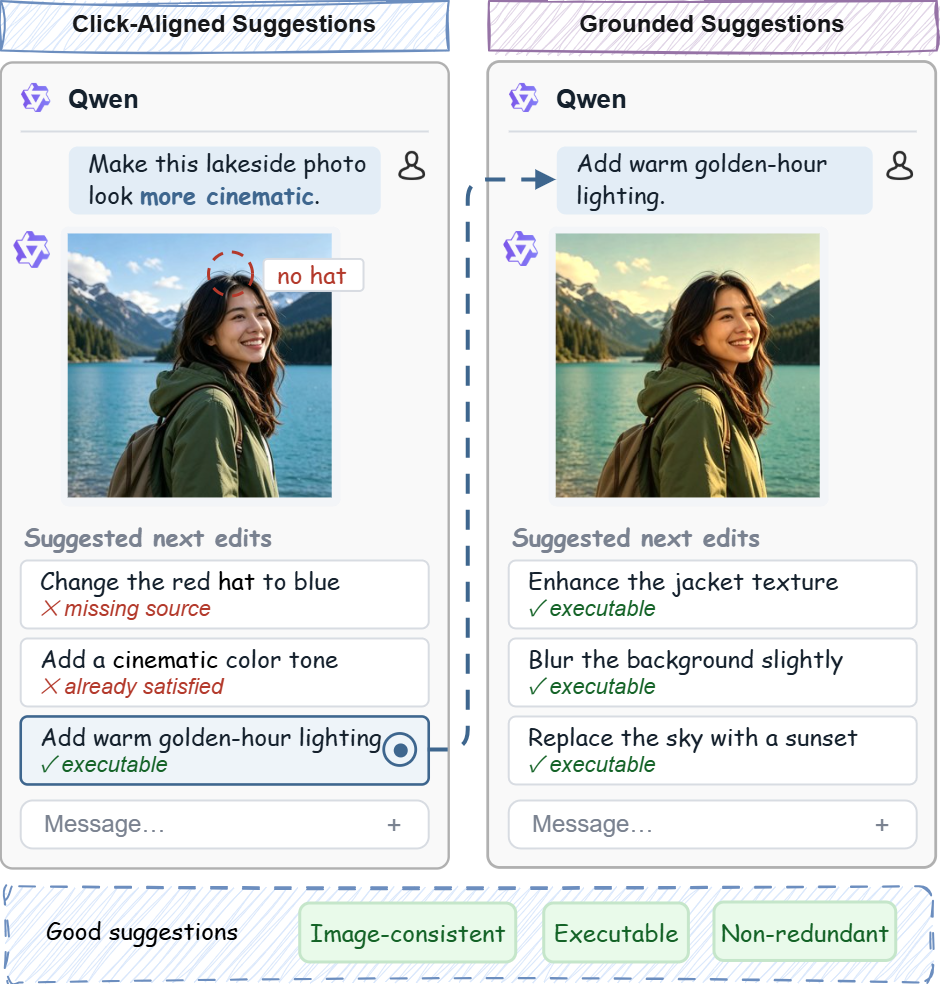}
\caption{Illustrative multi-turn policy comparison. A click-supervised policy without visual-consistency supervision (left) proposes a slate containing a missing-source edit and an already-satisfied target; the user selects the executable lighting edit. On the updated image, the full framework (right) produces executable, image-consistent, and non-redundant follow-up edits.}
\Description{Two panels compare suggestion slates for the same image and query. The upper panel includes an already-satisfied-target edit and a missing-source edit. The lower panel shows three executable suggestions after visual consistency checks.}
\label{fig:product}
\end{wrapfigure}

Conversational assistants increasingly recommend follow-up edits to help users continue multi-step tasks. Such suggestions play an important role in sustaining user engagement, improving retention, and enhancing user satisfaction. However, existing follow-up edit recommendation systems primarily focus on text-based conversations, leaving visual-creation conversations underexplored. In these conversations, users iteratively generate, edit, inspect, and refine images. A useful follow-up edit recommendation must therefore account for the latest visual state. After each round, a recommendation model maps the latest image, current query, and editing intent to a slate of follow-up edit suggestions (Figure~\ref{fig:product}). A useful slate should contain suggestions that users want to select, offer distinct creative directions, and remain executable on the current image.

To quantify the importance of visual context in this setting, we audit 100,000 adjacent user-turn pairs from real image-creation conversations in Qwen App. We find that 80.1\% of the follow-up editing queries are image-dependent: their intended edits depend on visual content that cannot be inferred from the preceding query alone and therefore require grounding in the latest image. Only 19.9\% are text-dependent, being supported by the preceding query or expressible through a generic edit template (Appendix~\ref{app:data}). Existing work addresses related but separate aspects of this setting. Query-suggestion systems learn user preferences from behavioral feedback but operate primarily on text \cite{cao2008context,sordoni2015hred,cici2025clicks,gqs2025}. Instruction-guided image editing executes user-specified edits \cite{brooks2023instructpix2pix}; image-editing recommendation generates diverse candidate instructions from an image and an underspecified prompt \cite{shen2024creativity}; and agentic editing decomposes a supplied editing goal into iterative actions \cite{zeng2026mira}. None jointly learns a behaviorally aligned follow-up slate and verifies its validity against the latest image. Multimodal follow-up edit recommendation must bridge this gap.

\begin{figure*}[t]
\centering
\includegraphics[width=\textwidth]{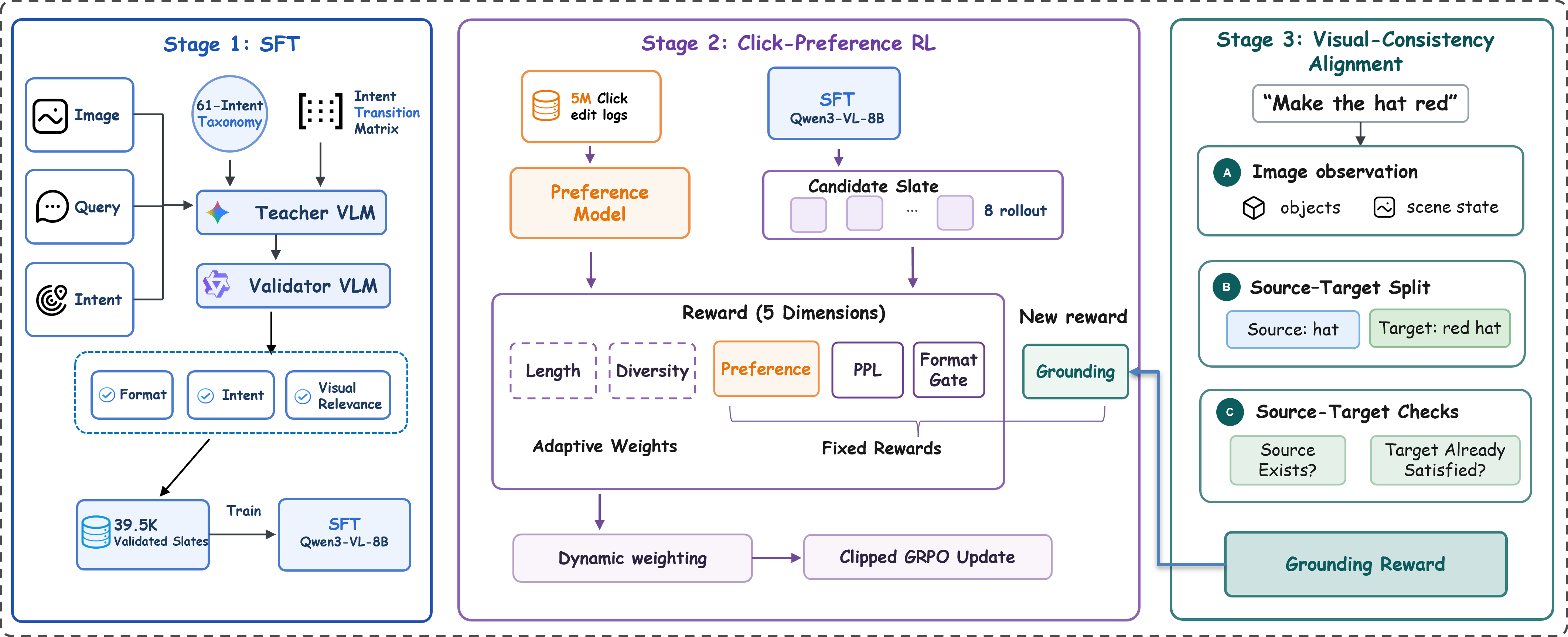}
\caption{Three-stage training framework. Stage 1 constructs validated SFT slates from real contexts. Stage 2 trains a click reward model and optimizes five non-grounding rewards with GRPO. Stage 3 derives an image-first source--target grounding reward and supplies it as a sixth signal to a separate GRPO run, initialized from SFT and otherwise using the Stage 2 pipeline.}
\Description{Three panels depict the training pipeline. Stage 1 maps an image, query, and intent through a 61-intent taxonomy, teacher VLM, validator, and format, intent, and visual-relevance checks to train an SFT policy on 39.5K validated slates. Stage 2 trains a preference model from 5M edit logs and optimizes five non-grounding rewards over eight policy rollouts before a clipped GRPO update. Stage 3 observes the image, splits a candidate into source and target, checks source existence and whether the target is already satisfied, and adds the grounding reward as the sixth signal.}
\label{fig:framework}
\end{figure*}

To bridge this gap, we present a three-stage framework (Figure~\ref{fig:framework}) that progressively incorporates three complementary sources of supervision: human-reviewed follow-up editing intents, behavioral preferences from real user feedback, and visual consistency with the current image. Stage 1 first addresses the lack of task-specific supervision: each real online training instance provides the latest image, current query, and editing intent, but does not contain a target slate of follow-up edit suggestions for SFT. It fills this supervision gap by combining these inputs with a human-reviewed table of appropriate follow-up editing intents. A vision-language teacher generates candidates for the allowed next intents, and the data pipeline validates and forms six-suggestion SFT targets. We then fine-tune the multimodal policy on these targets, establishing the task and its human-specified follow-up edit action space.

Stage 1 teaches the policy to generate task-appropriate follow-up suggestions within the human-defined action space, but provides no supervision about which suggestions users actually prefer. \textbf{Stage 2} therefore introduces behavioral supervision from real user clicks. To reduce display-position bias, we pair each clicked suggestion only with unclicked suggestions displayed above it. The resulting position-aware preference pairs are used to train an 8B vision-language reward model with the Bradley--Terry objective. Multi-objective Group Relative Policy Optimization (GRPO)~\cite{grpo2024} then optimizes the policy using the learned click-preference reward together with four complementary quality signals: format validity, distributional proximity to the SFT policy, content-aware length, and within-slate diversity.

However, these objectives provide no explicit supervision for visual consistency. Although Stage 2 improves expert-rated suggestion quality, the visual-inconsistency rate increases from 3.0\% after SFT to 3.7\% after click-based optimization (Table~\ref{tab:main}). Such inconsistencies arise when a suggestion relies on an absent source or requests a target state that is already satisfied. While the click reward model receives the image as input, its supervision comes solely from user choices, which reflect suggestion appeal rather than executability on the current image; the other four rewards constrain slate quality without directly evaluating visual consistency. \textbf{Stage 3} therefore introduces an image-first structured verifier. The verifier records the visual scene before reading the candidates, separates each suggestion into required sources and a target state, and independently checks whether each source exists and whether the target state is already satisfied. Its grounding score is added to GRPO as a sixth reward dimension. Because the verifier is used only during training, the deployed system retains a single 8B policy without additional serving latency.

In offline evaluation, the complete three-stage framework reduces visual inconsistency from 3.7\% to 0.9\% relative to Stage 2, while preserving expert-rated suggestion quality. We further deploy the framework in Qwen App and conduct a 14-day user-randomized A/B test involving millions of users. Relative to the previously deployed prompt-engineered (PE) policy, the framework improves recommendation CTR by 32.70\%, image take-away rate by 16.32\%, and average conversation turns per user by 39.90\%; all three lifts are statistically significant ($p<0.05$).

Taken together, our work makes the following contributions:
\begin{itemize}[leftmargin=*,topsep=2pt,itemsep=1pt]
\item We formulate multimodal follow-up edit recommendation for visual-creation conversations, where suggestions must align with user preferences while remaining valid for the current image. An audit of 100,000 follow-up editing queries and their preceding queries shows that 80.1\% depend on visual context.
\item We develop an end-to-end learning pipeline that constructs SFT targets from real user contexts using human-reviewed follow-up intents and validated teacher outputs, and then aligns the policy with actual user choices through position-aware click pairs and multi-objective GRPO.
\item We introduce an image-first source--target verifier that provides explicit visual-consistency supervision during RL by checking whether each suggestion relies on an absent source or requests an already-satisfied target state. It reduces visual inconsistency from 3.7\% to 0.9\% without degrading expert-rated suggestion quality.
\item We deploy the framework in Qwen App. A 14-day user-randomized A/B test involving millions of users demonstrates statistically significant improvements in recommendation CTR, image take-away rate, and average conversation turns per user.
\end{itemize}

\section{Related Work}
\label{sec:related}

\paragraph{Query and edit suggestion.}
Classical query suggestion retrieves or ranks candidates using co-occurrence, session graphs, and neural sequence models \cite{cao2008context,boldi2008queryflow,sordoni2015hred}. Conversational systems extend this setting by proactively generating suggestions that guide a multi-turn interaction \cite{rosset2020leading,gqr2025}. Recent deployed variants further model list diversity, cross-turn intent memory, retrieval context, or personalized openers \cite{relist2026,onepred2026,kuaishou2026,icebreaker2026}. Multimodal query suggestion has also been studied for image search \cite{wang2024mmqs}. More closely related to our task, image-editing recommendation generates diverse creative instructions from an image and an underspecified user prompt \cite{shen2024creativity}, whereas agentic editing decomposes and executes a supplied editing goal through iterative visual feedback \cite{zeng2026mira}. Our setting instead ranks follow-up edits after each completed edit and must align the slate with both behavioral preference and the updated visual state.

\paragraph{Click-based preference learning and position bias.}
Preference alignment commonly combines supervised fine-tuning, reward modeling, and policy optimization \cite{ouyang2022instructgpt}. For generative query suggestion, recent work turns clicks into preference signals for reward modeling or reinforcement learning \cite{cici2025clicks,gqs2025}, often alongside diversity-aware list objectives \cite{relist2026}. Clicks, however, reflect both preference and which suggestions users are likely to see \cite{joachims2002optimizing,craswell2008experimental}. Our position-aware construction pairs a clicked suggestion only with unclicked suggestions displayed above it, which the user was more likely to have seen \cite{joachims2005accurately}. This simple rule reduces, but does not remove, display-position bias; full propensity correction estimates viewing probabilities explicitly \cite{joachims2017unbiased,wang2018position,ai2018unbiased}. More fundamentally, even unbiased click preference does not determine whether a follow-up edit is visually consistent with the current image.

\paragraph{Visual grounding and hallucination alignment.}
Vision-language models can follow linguistic priors over conflicting pixels \cite{goyal2017vqa,agrawal2018vqacp}, a failure measured by object-hallucination metrics \cite{rohrbach2018object,pope2023} and suites for entangled illusions and unsupported assumptions \cite{hallusionbench2024,sesame2024}. Existing remedies either contrast visual and language-prior distributions at decoding time \cite{vcd2024} or align models with corrective or hallucination-aware preference feedback \cite{rlhfv2024,rlaifv2024,hadpo2023}. These approaches primarily assess whether a statement about an image is true. Follow-up edit recommendation instead requires visual edit validity: an edit may introduce a new target, but any source it consumes must be visible in the current image. This source--target asymmetry is not captured by caption-level factuality, motivating image-first verification of edit conditions.

\paragraph{Multi-objective reward optimization.}
Composite rewards help preference optimization satisfy multiple product requirements \cite{grpo2024,odin2024,dqo2026}. Because reward scales can differ, prior work motivates per-dimension normalization and adaptive weighting \cite{gdpo2026,dvao2026,mogrpo2025,dynweight2026,ctwa2026}. In our setting, these mechanisms balance behavioral preference and list-quality constraints, while the Stage 3 grounding reward separately optimizes visual consistency rather than click preference alone.

\section{Method}
\label{sec:method}

\subsection{Problem Formulation}
\label{subsec:formulation}

For each editing round, the input is $x=(I,q,e)$: the latest image, current query, and editing intent. The current query is rewritten from the multi-turn dialogue to incorporate relevant prior context; the policy and the Part~I teacher consume this rewritten query rather than the raw earlier turns. The policy $\pi_\theta$ generates an ordered candidate slate $Y=(y^1,\ldots,y^N)$, targeting six suggestions while accepting $5\leq N\leq7$ as product-valid. The serving layer randomly selects three distinct suggestions, indexed by $D$, to form the displayed slate $Y_D$. An RL sample of $Y$ is a rollout.

Our objective is to maximize the expected utility of the displayed slate:
\begin{equation}
\max_\theta\;
\mathbb E_{\substack{x\sim\mathcal X,\,
Y\sim\pi_\theta(\cdot\mid x),\\
D\sim\mathcal U_3(Y)}}
\left[U(Y_D\mid x)\right].
\label{eq:objective}
\end{equation}
Here $\mathcal U_3(Y)$ denotes random selection of three distinct suggestions from $Y$. A useful displayed slate should attract user clicks while remaining well formed, diverse, and valid for the current image. Parts~I--III introduce the supervision and rewards used to optimize these properties.

\subsection{System Overview}
\label{subsec:overview}

The framework adds supervision in three stages (Figure~\ref{fig:framework}). \textbf{Stage 1} builds six-suggestion SFT targets from real inputs, a human-reviewed table of appropriate follow-up editing intents, and validated teacher outputs. \textbf{Stage 2} trains a vision-language RM from position-aware click pairs and optimizes five complementary rewards with GRPO. \textbf{Stage 3} uses a structured verifier to check required sources and target states, adding a grounding reward to optimize visual consistency. The verifier is used during Stage 3 training and checkpoint selection, but not at serving time; deployment uses a single 8B policy followed by the existing random display selection. Part~II occupies two panels in the figure because it includes both RM training and GRPO.

\subsection{Part I: SFT Data Construction from Real Online Contexts}
\label{sec:part1}

\begin{figure*}[t]
\centering
\includegraphics[width=\textwidth]{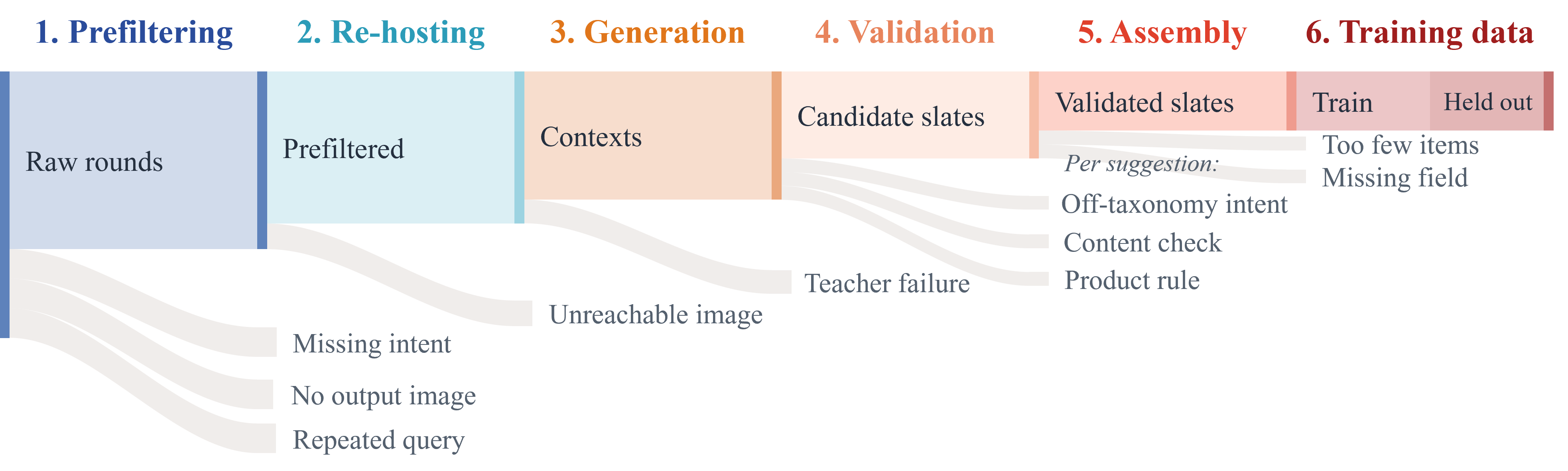}
\caption{SFT data-construction pipeline (general track). Pale blocks are the data surviving each stage; grey ribbons show removed records and their filtering reasons.}
\Description{A left-to-right pipeline diagram with six stages in which the surviving data narrows at each stage, with labeled grey branches showing why records are dropped.}
\label{fig:datapipe}
\end{figure*}

Part~I starts before the product has follow-up edit slates or clicks on such slates, so there is no recommendation target to copy. We only have the real task inputs $(I,q,e)$. We combine them with a human-reviewed table of appropriate follow-up editing intents, ask a teacher to write candidates, and then validate and form the candidates into SFT targets. Figure~\ref{fig:datapipe} shows this pipeline.

\subsubsection{From real online inputs to editing contexts}
\label{subsec:logs}
Each sampled request provides the latest output image, rewritten current query, and parsed intent. We remove requests without a usable image and merge consecutive copies of the same query, which usually come from retries. Because the original image URLs later become unavailable, we copy each retained image to stable storage. This leaves 44.3K usable contexts from 120.0K requests. The rewritten query incorporates relevant multi-turn context, so neither the teacher nor the deployed policy needs the raw earlier turns.

\subsubsection{Intent-guided generation}
\label{subsec:gen}
We define 61 editing intents and manually construct a table of appropriate follow-up editing intents for each current intent. Human reviewers check this intent table before generation. We use three generation tracks: a general track for all contexts, a priority track for common high-traffic cases, and a continuation track that fills incomplete slates with additional validated suggestions (Appendix~\ref{app:gentracks}). Gemini 3 Flash~\cite{gemini3flash2025} receives only $(I,q,e)$ and the allowed follow-up intents. It writes raw candidates grouped by follow-up intent. In short, online inputs provide the context, people define the allowed directions, and the teacher turns those directions into candidate text.

\subsubsection{Validation, assembly, and fine-tuning}
\label{subsec:validate}
Validation is applied to each suggestion. An intent-label check removes candidates assigned to an invalid intent. A Stage 1 validator checks structure, wording, repetition, and coarse image relevance. A separate product-rule layer removes restricted content. These checks reduce the general track from 359K to 243K suggestions. If fewer than six remain, we refill only with validated continuation candidates. We keep exactly six suggestions in predefined intent order and drop contexts that still cannot be completed, producing 41.6K full slates and 39.5K training slates. Qwen3-VL-8B~\cite{qwen3vl2025} is then fine-tuned with rank-4 LoRA~\cite{hu2022lora} and a frozen visual encoder. This SFT model initializes the RL actor and serves as the reference policy.

\subsection{Part II: Click-Based Reward Modeling and Multi-Objective RL}
\label{sec:part2}

SFT teaches the policy to generate well-formed suggestions that follow the human-reviewed intent table. However, online feedback reveals a clear gap between these rule-guided SFT suggestions and what users actually prefer: an edit can be appropriate under the intent table without being the option that users are most willing to select. Further imitation of the same rule-guided targets does not directly use this choice signal. We therefore learn a preference reward model from real clicks and use GRPO to optimize the policy toward actual user choices.

\subsubsection{Position-Aware Behavioral Preference}
\label{subsec:preference}

The production generator first creates a candidate slate, from which the display layer randomly selects three suggestions. Consequently, one logged impression contains the current image, query, intent, the three displayed suggestions, and the clicked suggestion. We treat all turns within one user conversation as a session and keep at most one eligible impression from each non-empty session. We retain impressions with a valid click and at least two usable suggestions, and remove invalid or generic options.

Because clicks depend on display position \cite{craswell2008experimental}, we only compare a clicked suggestion with unclicked suggestions shown above it, which the user was likely to have seen. The clicked suggestion is $y^+$, and each unclicked usable suggestion above it becomes a separate $y^-$. Suggestions below the click are excluded because they may not have been viewed. This position-aware rule reduces the influence of display bias, although it cannot remove it completely.

This construction produces 173{,}071 pairs, split by request ID into 164{,}401 training pairs and 8{,}670 validation pairs. Appendix~\ref{app:data} gives the complete data funnel.

For context $x$, clicked suggestion $y^+$, and an unclicked suggestion $y^-$ displayed above it, we train an 8B vision-language RM with the Bradley--Terry objective \cite{bradley1952rank,ouyang2022instructgpt}:
\begin{equation}
\mathcal L_{\mathrm{RM}}
=-\mathbb E\log\sigma\!\left(r_\phi(x,y^+)-r_\phi(x,y^-)\right).
\label{eq:bt}
\end{equation}
The RM receives the image, query, intent, and one suggestion and returns a scalar click-preference score. To score a candidate slate, we average its suggestion scores:
\begin{equation}
\bar r_\phi(Y)=\frac{1}{N}\sum_{j=1}^{N}r_\phi(x,y^j),\qquad 5\leq N\leq7.
\label{eq:prefaggregate}
\end{equation}
Averaging suggestion scores also matches their expected mean under random three-suggestion display.

\subsubsection{Five Rewards for Stage 2}
\label{subsec:rewards}

The preference score alone is easy to exploit: the policy can produce malformed, long, or repetitive text that the RM still likes. Stage 2 therefore optimizes five parallel signals: click preference, output validity, perplexity (PPL), content-aware length, and within-list diversity. The following definitions specify their roles and aggregation levels; higher is better for all five rewards.
\par\textbf{Click Preference Reward.} The suggestion-level RM measures how likely a suggestion is to be selected. Equation~\ref{eq:prefaggregate} averages the six suggestion scores into the slate-level reward $\bar r_\phi(Y)$.

\par\textbf{Output Gate Reward.} The slate-level binary reward $r_{\mathrm{gate}}$ is one when the output is valid JSON, contains 5--7 parsed suggestions, and uses only allowed intent labels; otherwise it is zero. This range is the product-valid contract; generation is still prompted and supervised toward the candidate-slate target of six.

\par\textbf{Perplexity (PPL) Reward.} RL can sample lists that are unlikely under the SFT policy. We therefore use the mean log-likelihood under the fixed SFT policy, equivalently negative log perplexity $-\log(\mathrm{PPL})$, as a rollout-level reward that keeps generation close to the SFT distribution:
\begin{equation}
r_{\mathrm{ppl}}(Y)=\frac{1}{T}\sum_{t=1}^{T}
\log\pi_{\mathrm{SFT}}(Y_t\mid Y_{<t},x),
\label{eq:ppl}
\end{equation}
Thus, a rollout that is more familiar to the SFT policy receives a higher score.

\par\textbf{Content-Aware Length Reward.} The suggestion-level reward $r_{\mathrm{len}}$ uses a character budget matched to the product display. Let $\ell_j$ be the character length and $d_j$ the content-density class:
\begin{align}
z_j&=\frac{L_2(d_j)-\ell_j}{L_2(d_j)-L_1(d_j)},\\
r_{\mathrm{len}}(y^j)&=\operatorname{clip}(z_j,0,1).
\label{eq:length}
\end{align}
The score is one at or below $L_1(d_j)$ and decreases linearly to zero at $L_2(d_j)$. More specific suggestions, such as edits that name an object or region, receive a larger length budget, and the six suggestion scores are averaged. The content-density class is produced by the same structured image analysis used by the verifier.

\par\textbf{Within-List Diversity Reward.} For semantic diversity, Qwen3-Embedding \cite{qwen3embedding2025} cosine similarities are calibrated to $\widetilde s_{jj'}\in[0,1]$. The slate reward is
\begin{equation}
r_{\mathrm{div}}(Y)=1-\max_{j<j'}\widetilde s_{jj'}.
\label{eq:div}
\end{equation}
Using the most similar pair makes one near-duplicate visible; a mean over all 15 pairs can hide it. Appendix~\ref{app:optimization} gives the gate handling, length thresholds, and other implementation constants.

\subsubsection{Multi-Objective GRPO}
\label{subsec:grpo}

GRPO uses 98K training contexts. For each context, the policy samples $G=8$ slates. The five reward dimensions are indexed by
\begin{equation}
\mathcal Q_5=\{\mathrm{gate},\phi,\mathrm{ppl},
\mathrm{len},\mathrm{div}\}.
\end{equation}
Let $r_i^k$ be reward $k$ for rollout $i$. For each $k\in\mathcal Q_5$, we compute the mean $\mu_{k,x}$ and sample standard deviation $\sigma_{k,x}$ across the eight rollouts:
\begin{equation}
\widetilde A_i^k=
\frac{r_i^k-\mu_{k,x}}{\sigma_{k,x}+\epsilon_\sigma}.
\label{eq:dimnorm}
\end{equation}
If all eight scores are equal, this dimension contributes zero. Normalizing each reward separately prevents the large numerical scale of one reward from hiding the others. We then take a weighted sum and apply masked batch standardization \cite{grpo2024}:
\begin{equation}
A_i=\operatorname{Standardize}_{\mathcal B}\!\left(
\sum_{k\in\mathcal Q_5}\lambda_k\widetilde A_i^k
\right).
\label{eq:advantage}
\end{equation}

\par\textbf{Dynamic Text-Side Weighting.} Only the length and diversity weights change during training. After each batch, $m_{\mathrm{len}}$ is the mean length reward and $m_{\mathrm{div}}$ is the fraction of slates below the diversity floor. We smooth each metric with an exponential moving average $\widetilde m_k$. The signed gap $\delta_k$ is positive when the corresponding length or diversity target is missed:
\begin{equation}
\begin{aligned}
\widetilde m_k&\leftarrow(1-\eta)\widetilde m_k+\eta m_k,\qquad
\lambda_k&\leftarrow\operatorname{clip}(\lambda_k+\kappa\delta_k,0,\lambda_{\max}).
\end{aligned}
\label{eq:textcontroller}
\end{equation}
The weight rises when the constraint is missed and falls when it is met. We cap it because too much diversity pressure can make the model invent unrelated objects merely to make the six suggestions look different. Gate, preference, and PPL stay fixed. Appendix~\ref{app:optimization} gives the targets, update rate, and all initial weights.

We optimize $A_i$ with the standard clipped GRPO objective and entropy bonus \cite{grpo2024}, together with KL regularization toward the SFT policy \cite{ouyang2022instructgpt}. The PPL reward scores each sampled slate under the fixed SFT model, whereas KL constrains the policy distribution during optimization. Appendix~\ref{app:training} gives the optimizer and coefficients.

\subsection{Part III: Visual Consistency Optimization}
\label{sec:part3}

Evaluation reveals that Stage 2 improves expert-rated quality but raises visual inconsistency from 3.0\% for SFT to 3.7\%. This regression suggests that optimizing user preferences alone may favor appealing or creative suggestions without ensuring that they remain consistent with the current image.

The click RM can perceive the image, but it is trained only with user-preference labels. The other rewards control generation and list quality rather than visual consistency. Part~III therefore adds direct image-side supervision: a structured verifier checks the visual conditions required by each edit and provides the sixth reward.

Throughout Part~III, visual consistency denotes the desired policy property, whereas grounding denotes the operational reward and audit signals used to optimize and measure it.

\subsubsection{Defining Visual Consistency}
\label{subsec:definition}

An edit instruction describes a change from the current image to a new state. A source is an object or state that must already be present in the current image. A target is the desired state after editing. We split each candidate $y$ into
\begin{equation}
y\;\longrightarrow\;
\begin{cases}
\mathcal S(y): & \text{required sources},\\
\mathcal T(y): & \text{target state},
\end{cases}
\label{eq:parse}
\end{equation}
where $\mathcal S(y)$ contains the required sources and $\mathcal T(y)$ is the target state. The two parts follow opposite rules: every source must already be visible, whereas a visually checkable target should not already be satisfied.

\begin{wrapfigure}{r}{0.48\textwidth}
\centering
\includegraphics[width=\linewidth]{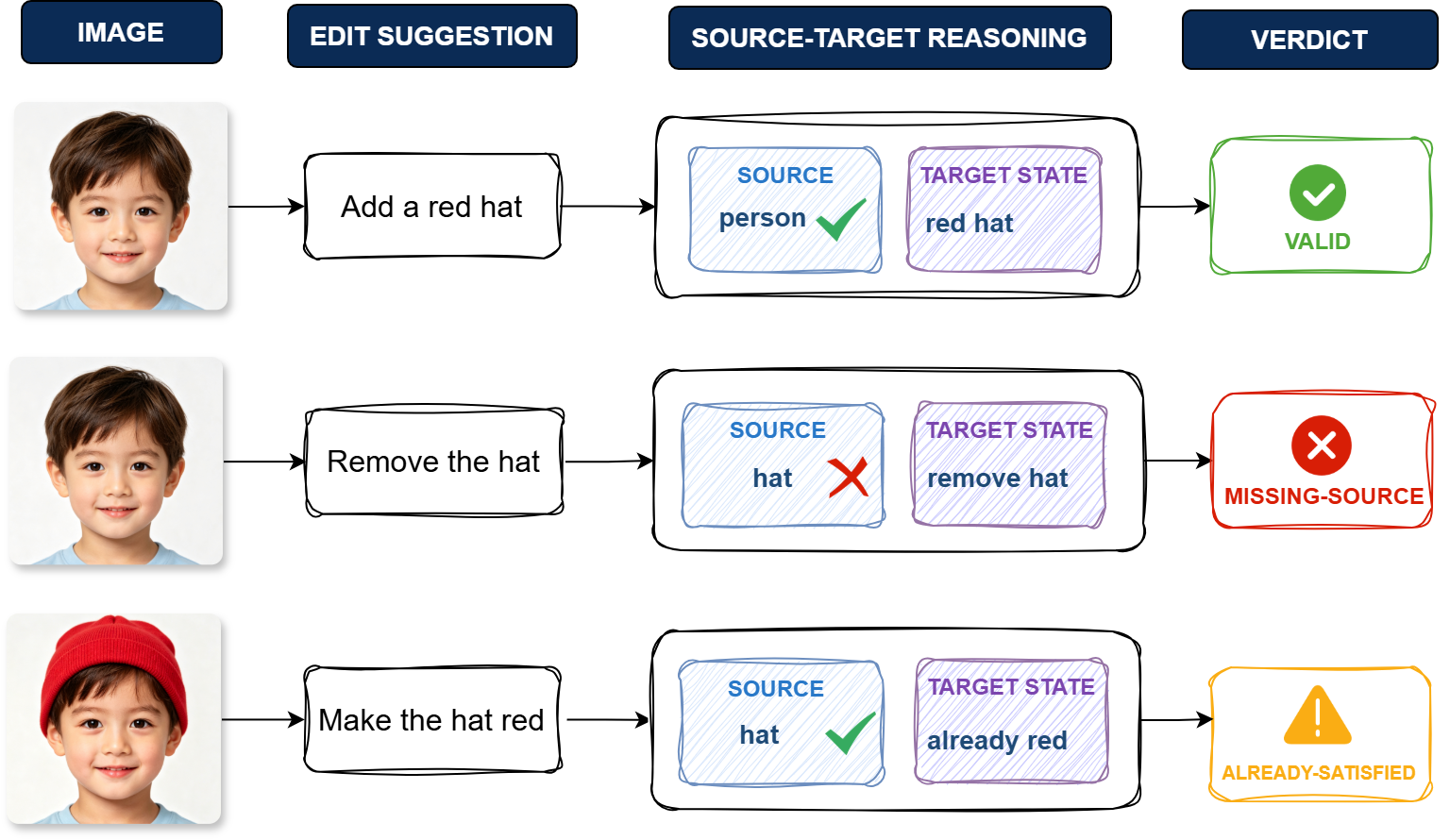}
\caption{Source--target visual consistency checks. A suggestion is invalid when it presupposes an absent source or requests a visually checkable target state that already holds.}
\Description{Three examples illustrate source--target visual consistency checks. Adding a red hat is valid because the person exists and the new hat is a target. Removing a nonexistent hat is a missing-source error. Making an already red hat red is an already-satisfied-target error.}
\label{fig:source-target}
\end{wrapfigure}

For example, ``remove the hat'' needs a visible hat. ``Add a hat'' treats the hat as a target and does not require it beforehand. ``Make the hat red'' needs a hat whose current state is not already red. Figure~\ref{fig:source-target} works through these three cases.

A candidate has a visual inconsistency when either requirement breaks. This gives two failure modes:
\begin{equation}
\begin{aligned}
v_{\mathrm{ms}}(y,I)&=\mathbf{1}\big[\text{some }s\in\mathcal S(y)\text{ is absent from }I\big],\\
v_{\mathrm{as}}(y,I)&=\mathbf{1}\big[\mathcal T(y)\text{ already holds in }I\big],\\
v(y,I)&=v_{\mathrm{ms}}(y,I)\lor v_{\mathrm{as}}(y,I),
\end{aligned}
\label{eq:inconsistency}
\end{equation}
A missing-source error occurs when an edit requires an object or state that is absent from the current image. An already-satisfied-target error occurs when the requested post-edit state is already present, making the edit redundant. A candidate is valid when neither occurs.

We keep the two flags separate because finding an object and comparing a state are different visual tasks. If a target is subjective or cannot be judged from the image, the target check is left uncertain and fails open rather than flagging an error. During training, the verifier supplies the corresponding estimates $\widehat v_{\mathrm{ms}}$, $\widehat v_{\mathrm{as}}$, and $\widehat v$.

\subsubsection{Image-First Source--Target Verification}
\label{subsec:verifier}

Prior work shows that vision-language models can rely on linguistic priors in a question instead of the visual evidence \cite{goyal2017vqa,agrawal2018vqacp}, hallucinate objects that are not present \cite{pope2023}, and accept unsupported assumptions introduced by the prompt \cite{hallusionbench2024,sesame2024}. In our task, the candidate itself can introduce such an assumption. A one-pass verifier may assume that a hat exists after reading ``remove the hat,'' or reject ``add a hat'' simply because the new hat is not visible yet. We therefore use a frozen Qwen3-VL-30B-A3B verifier \cite{qwen3vl2025} and split verification into four explicit steps.

\par\textbf{Image Observation.} Before reading any candidate, the verifier records visible objects, people, text, regions, and a compact scene state. This prevents candidate wording from changing the initial image description.

\par\textbf{Source--Target Split.} The verifier then records the operation, required sources, and target state. A source must be quoted from the candidate, and an attribute such as color, pose, or style cannot be a source by itself. Appendix~\ref{app:verifier} lists the operation rules.

\par\textbf{Source--Target Checks.} Source existence and whether the target state is already satisfied are checked separately. A source counts as present only when the verifier gives a short location and appearance description. For the target check, it writes the desired and current states side by side. Content that the edit intends to add is never required to exist beforehand, and subjective targets fail open.

\par\textbf{Visual Inconsistency Detection.} The verifier emits source-existence fields and a target-state comparison in its JSON output; Appendix~\ref{app:verifier} gives the full schema. These structured fields are treated as the authoritative outputs, and the final inconsistency flags are derived from them rather than from a free-form explanation. After parsing, one-directional guards may clear a small set of known parser-induced false alarms, but they never introduce a new inconsistency. Uncertain, unparseable, or failed calls are treated as having no flagged inconsistency during training and are logged for monitoring.

\subsubsection{Grounding as the Sixth Reward}
\label{subsec:grounding}

One deterministic verifier call scores all suggestions in a rollout. For a realized slate $Y$ with $N$ parsed suggestions, the grounding reward is
\begin{equation}
\widehat g(Y,I)=1-\frac{1}{N}\sum_{j=1}^{N}\widehat v(y^j,I).
\label{eq:ground}
\end{equation}
Averaging over suggestions penalizes a bad suggestion without rejecting the rest of the slate. An uncertain or unparseable suggestion uses $\widehat v=0$, a fail-open choice that avoids noisy penalties but can miss real errors. Higher $\widehat g$ is better.

Grounding follows the same per-reward group normalization as the other rewards. It is computed over structurally valid rollouts; gate-failed rollouts receive zero grounding advantage. The full advantage is
\begin{equation}
A_i=\operatorname{Standardize}_{\mathcal B}\!\left(
\sum_{k\in\mathcal Q_6}\lambda_k\widetilde A_i^k
\right),
\label{eq:advantagefull}
\end{equation}
where $\mathcal Q_6=\mathcal Q_5\cup\{\mathrm{grd}\}$ and $\lambda_{\mathrm{grd}}=0.5$ is fixed. Stage 2 removes only grounding. Appendix~\ref{app:optimization} gives the masked normalization details.

\section{Experiments}
\label{sec:eval}

We evaluate three questions: whether the full framework improves end-to-end offline quality and online engagement, which reward components contribute to these outcomes, and whether source--target structure is necessary for reliable verification.

\subsection{Experimental Setup and Metrics}

\par\textbf{Evaluation Protocol and Policy Variants.} We evaluate PE and the learned policies on 500 disjoint real Qwen App sessions, using identical decoding and four generated slates per session for every arm. One evaluation unit is one generated slate, and every automated comparison scores all checkpoints on the same units. The stages are cumulative: Stage 1 is SFT; Stage 2 adds click-based multi-objective RL; and Stage 3 adds the visual verifier's grounding reward. PE is a prompt-engineered instantiation of the same Qwen3-VL-8B backbone: it receives the same latest image, query, and intent; uses the same candidate-slate protocol, decoding, and shared display layer; but has no SFT, click-RL, or grounding-reward update. It is the reference for GSB and all online metrics. A separate calibration set evaluates the verifier itself. Appendix~\ref{app:data} gives the denominators, and Section~\ref{subsec:verifierstudy} describes the verifier benchmark.

\par\textbf{Offline Metrics.} We use an external grounding audit to measure visual inconsistency and GSB to measure expert-rated suggestion quality. Gemini 3.1 Pro \cite{gemini31pro2026}, from a different model family and provider than the Qwen verifier used in training, evaluates the audit. This separation avoids letting the training verifier judge outputs shaped by its own reward. Because visual inconsistencies are rare, a stratified blind expert review of 1{,}276 suggestions validates the Gemini evaluation; it reaches 90.7\% recall and 89.0\% precision on the inconsistency class (Appendix~\ref{app:groundaudit}). Eight experts score randomized outputs from all arms, including PE, on a 0--3 scale without seeing policy identities or reward values; every arm receives the same 800 suggestion judgments. GSB is the aggregate expert-score difference from PE under this common protocol, so the PE row is the reporting reference at zero rather than an unscored quality value. Appendix~\ref{app:human} gives the rubric, formula, and quality-control procedure.

\par\textbf{Online A/B Test.} The 14-day experiment runs all policy arms concurrently on the same eligible Qwen App population, with user-level randomization. Each arm receives the same 5\% traffic allocation and shares the same display logic: each policy generates a candidate slate, from which the display layer randomly selects three suggestions. PE runs concurrently as the common control arm. Recommendation CTR measures selection from the three displayed edits, image take-away rate measures whether users keep the edited image, and turns per user measures continued editing. We report PE-relative lifts; all reported lifts are significant at $p<0.05$. Latency, generation failure, and negative feedback are monitored as launch constraints. Appendix~\ref{app:online} gives the full setup.

\par\textbf{Implementation Details.} PE, the actor, the SFT reference, and the click RM use Qwen3-VL-8B; GRPO samples eight rollout slates per context. The training verifier, diversity encoder, and remaining training configuration are specified in Appendix~\ref{app:training}.

\subsection{End-to-End Results}

Table~\ref{tab:main} combines offline diagnostics and online outcomes. PE runs concurrently as the online control and is the GSB base; its 8.6\% visual inconsistency and 23.3\% redundancy are measured with the same offline protocol. All policy arms use matched traffic allocation and display logic, so their PE-relative lifts compare policies under matched serving conditions. Expert GSB rises from $+332$ for SFT to $+405$ for Stage 2 and $+446$ for the full framework. Visual inconsistency does not follow that order: Stage 2 raises the rate from 3.0\% to 3.7\%, whereas the full framework reduces it to 0.9\% and lowers redundancy to 8.8\%.

SFT improves all three online metrics over PE. Stage 2 achieves the largest CTR lift, but its take-away rate and turns per user fall below SFT. The full framework retains a similar CTR lift while achieving the best take-away rate and longest conversations. Together with the visual-consistency audit, this pattern suggests that the full framework improves the subsequent editing path rather than clicks alone.

Low visual inconsistency is meaningful only if quality and list variety are preserved. The full framework improves expert quality and lowers within-list redundancy to 8.8\%. Appendix~\ref{app:qualitative} shows paired examples.

\begin{table}[t]
\caption{Core end-to-end results. Top: offline metrics. Bottom: PE-relative online lift.}
\label{tab:main}
\reportfloatwidth{0.62}
\footnotesize
\begin{tabular*}{\columnwidth}{@{\extracolsep{\fill}}p{0.36\columnwidth}rrr@{}}
\toprule
\multicolumn{4}{@{}l@{}}{\textit{Offline evaluation}}\\
\cmidrule(r){1-4}
Policy & GSB $\uparrow$ & Ground. $\downarrow$ & Redund. $\downarrow$\\
\midrule
PE (base) & $+0$ & 8.6\% & 23.3\%\\
SFT (Stage 1) & $+332$ & 3.0\% & 17.2\%\\
SFT + RL (Stage 2) & $+405$ & 3.7\% & 11.9\%\\
Full framework (Stage 3) & $\mathbf{+446}$ & \textbf{0.9\%} & \textbf{8.8\%}\\
\midrule
\multicolumn{4}{@{}l@{}}{\textit{Online lift over PE}}\\
\cmidrule(r){1-4}
Policy & CTR $\uparrow$ & Take-away $\uparrow$ & Turns/user $\uparrow$\\
\midrule
PE (base) & 0.00\% & 0.00\% & 0.00\%\\
SFT (Stage 1) & +25.33\% & +13.70\% & +33.64\%\\
SFT + RL (Stage 2) & \textbf{+33.48\%} & +7.50\% & +32.56\%\\
Full framework (Stage 3) & +32.70\% & \textbf{+16.32\%} & \textbf{+39.90\%}\\
\bottomrule
\end{tabular*}
\end{table}

\subsection{Ablation Study 1: RL Components}
\label{subsec:ablation}

Table~\ref{tab:rewardablation} ablates RL components cumulatively, with rows in training order. Every row starts from the SFT policy. The first RL row adds click-preference optimization together with three fixed safeguards: a rule-based output gate, PPL, and content-aware length. The next rows add max-pair diversity, per-dimension normalization with dynamic text-side weighting, and finally the grounding reward.

\begin{table}[H]
\caption{Cumulative ablation of RL components. All rows start from the SFT policy.}
\label{tab:rewardablation}
\reportfloatwidth{0.60}
\footnotesize
\setlength{\tabcolsep}{2pt}
\begin{tabular*}{\columnwidth}{@{\extracolsep{\fill}}lccc@{}}
\toprule
RL configuration & GSB vs base $\uparrow$ & Ground. $\downarrow$ & Redund. $\downarrow$\\
\midrule
SFT initialization & $+332$ & 3.0 & 17.2\\
$+$ core RL rewards & $+374$ & 4.4 & 28.1\\
$+$ max-pair diversity & $+383$ & 4.0 & 19.1\\
$+$ norm./dynamic weights & $+405$ & 3.7 & 11.9\\
$+$ grounding reward & $\mathbf{+446}$ & \textbf{0.9} & \textbf{8.8}\\
\bottomrule
\end{tabular*}
\end{table}

Preference RL lengthens and repeats suggestions; these changes are associated with more visual inconsistencies and higher redundancy. Max-pair diversity repairs most of the list repetition (Appendix~\ref{app:diversity}); per-dimension normalization and dynamic text-side weighting further lower redundancy while recovering part of the visual regression. Adding the grounding reward produces the lowest observed inconsistency rate (0.9\%) and the highest GSB.

\subsection{Ablation Study 2: Click Pair Construction and Reward Model Scale}

Position-aware pair construction improves held-out click accuracy from 0.619 to 0.690 at a matched 51.9K budget and a shared 2B backbone (Section~\ref{subsec:preference}). This result isolates the benefit of respecting display position when deriving pairwise labels.

We next ask what click supervision can and cannot learn. On 2{,}199 within-request pairs, Table~\ref{tab:rm} reports agreement on clear pairs with an expert-tier gap of at least two, and the good-minus-bad gap between tiers 2--3 and tiers 0--1 in mean within-RM percentile. We also measure the residual inversion between the two lowest expert tiers. A well-ordered RM should score tier 0 below tier 1; a positive Tier0--Tier1 gap therefore indicates an undesirable inversion. Tier-0 failures include severe cases such as suggestions that rely on image content that is not present.

\begin{table}[H]
\caption{Expert-transfer reward-model study (2{,}199 within-request pairs; independently trained reward-model checkpoints).}
\label{tab:rm}
\reportfloatwidth{0.72}
\footnotesize
\setlength{\tabcolsep}{4pt}
\begin{tabular*}{\columnwidth}{@{\extracolsep{\fill}}lcccc@{}}
\toprule
Reward model & Pairs & Expert agr.\ $\uparrow$ & Good$-$Bad $\uparrow$ & Low-tier inv.\ $\downarrow$\\
\midrule
Raw-pair (2B) & 125K & 0.490 & $+0.015$ & 0.090\\
Position-aware (2B) & 160K & 0.515 & $+0.048$ & \textbf{0.040}\\
Position-aware (8B) & 160K & \textbf{0.569} & $\mathbf{+0.086}$ & 0.041\\
\bottomrule
\end{tabular*}
\end{table}

Better pairs and a larger backbone improve expert agreement and the good-minus-bad gap. However, the tier-0 versus tier-1 inversion halves with position-aware pairs and then stalls at 8B. Although the RM receives the image, its supervision still comes from clicks. A larger model therefore becomes better at estimating what users find appealing, including a measurable preference for longer, more elaborate phrasing (Appendix~\ref{app:lengthbias}), but does not reliably determine whether an edit is possible on the image.

Improved pair construction and model scale strengthen behavioral preference modeling, but neither supplies direct supervision for visual consistency. We therefore address the remaining gap with a separately supervised image-reading verifier rather than a still larger click RM.

\subsection{Ablation Study 3: Source--Target Structure in the Verifier}
\label{subsec:verifierstudy}

We compare the source--target verifier with a single-pass baseline that scores each candidate without image-first observation or a source--target split. Both use the same production rubric (Appendix~\ref{app:verifier}), isolating the contribution of structured verification.

On 488 visual inconsistencies and 491 excellent suggestions, source--target verification recalls 78.7\% and falsely rejects 0.6\%, against 47.5\% and 22.2\% for the single-pass baseline. Labels combine expert worksheets with semantically constructed production cases (Appendix~\ref{app:verifier}).

Table~\ref{tab:verifier} shows that source--target verification recalls 92.9\% of missing sources and 74.5\% of already-satisfied targets, against 61.6\% and 43.4\% for the single-pass baseline. It flags 4.0\% of poor suggestions that conflict with an explicit user requirement but are not visually inconsistent, compared with 60.0\% for the baseline. Separating required sources from targets an edit may create keeps the verifier focused on visual inconsistency.

Recall alone is insufficient for RL: a 22.2\% false-rejection rate would penalize valid creative edits and encourage safe, generic outputs. The source--target verifier's 0.6\% rate preserves that space while improving recall.

\begin{table}[H]
\caption{Verifier evaluation on the visual-consistency calibration set.}
\label{tab:verifier}
\reportfloatwidth{0.68}
\footnotesize
\setlength{\tabcolsep}{4pt}
\begin{tabular*}{\columnwidth}{@{\extracolsep{\fill}}lcccc@{}}
\toprule
& \multicolumn{3}{c}{Visual inconsistency recall $\uparrow$} & False\\
\cmidrule(lr){2-4}
Verifier & Missing src. & Already sat. & Union & rejection $\downarrow$\\
\midrule
Single-pass VLM & 61.6\% & 43.4\% & 47.5\% & 22.2\%\\
Source--target & \textbf{92.9\%} & \textbf{74.5\%} & \textbf{78.7\%} & \textbf{0.6\%}\\
\bottomrule
\end{tabular*}
\end{table}

\subsection{Stage-Wise Visual Inconsistency Analysis}

The verifier study above evaluates the signal itself; Table~\ref{tab:errortypes} evaluates its policy-level consequence when used as the Stage 3 reward. It decomposes visual inconsistency across the three training stages on the later offline evaluation set.

\begin{table}[H]
\caption{Stage-wise visual inconsistency by error type. Union denotes the rate of suggestions with either error.}
\label{tab:errortypes}
\reportfloatwidth{0.70}
\footnotesize
\setlength{\tabcolsep}{5pt}
\begin{tabular*}{\columnwidth}{@{\extracolsep{\fill}}p{0.32\columnwidth}rrr@{}}
\toprule
Policy & \shortstack{Missing\\source $\downarrow$} & \shortstack{Already-satisfied\\target $\downarrow$} & \shortstack{Visual inconsistency\\(union) $\downarrow$}\\
\midrule
SFT (Stage 1) & 0.74\% & 2.26\% & 3.0\%\\
SFT + RL (Stage 2) & 0.70\% & 3.00\% & 3.7\%\\
Full framework (Stage 3) & \textbf{0.42\%} & \textbf{0.48\%} & \textbf{0.9\%}\\
\bottomrule
\end{tabular*}
\end{table}

The three stage policies are evaluated on the same held-out sessions with identical decoding. Relative to Stage 2, Stage 3 reduces missing-source errors by 40.0\% (0.70\% to 0.42\%) and already-satisfied-target errors by 84.0\% (3.00\% to 0.48\%). Together, these changes yield a 75.7\% reduction in union visual inconsistency (3.7\% to 0.9\%). They show that the visual-consistency reward addresses both error types rather than only the easier missing-source case.

\section{Conclusion}

Follow-up edit recommendation in Qwen App requires both clicks and pixels: suggestions must reflect user preferences while remaining executable on the current image. Our three-stage framework constructs SFT targets from real contexts and a human-reviewed intent table, aligns the policy with position-aware click preferences through multi-objective RL, and adds image-first source--target verification as visual-consistency supervision. In a 14-day online study with 5\% traffic and millions of users per arm, the full framework delivers statistically significant ($p<0.05$) PE-relative lifts of 32.70\% in CTR, 16.32\% in image take-away rate, and 39.90\% in average conversation turns per user. It also reduces visual inconsistency from 3.7\% for Stage 2 to 0.9\% while preserving expert-rated quality. These results show the complementary value of behavioral and visual supervision for useful follow-up editing.

\newpage

\bibliographystyle{ACM-Reference-Format}
\bibliography{references}

@inproceedings{cici2025clicks,
  author    = {Yin, Junhao and Wang, Haolin and Bao, Peng and Xu, Ju and Wang, Yongliang},
  title     = {From Clicks to Preference: A Multi-stage Alignment Framework for Generative Query Suggestion in Conversational System},
  booktitle = {Proceedings of the ACM SIGKDD Conference on Knowledge Discovery and Data Mining (KDD)},
  year      = {2026},
  pages     = {2539--2550},
  doi       = {10.1145/3770854.3783953},
  publisher = {ACM},
  address   = {New York, NY, USA}
}

@inproceedings{gqs2025,
  author    = {Min, Erxue and Huang, Hsiu-Yuan and Yang, Xihong and Yang, Min and Jia, Xin and Wu, Yunfang and Cai, Hengyi and Wang, Junfeng and Wang, Shuaiqiang and Yin, Dawei},
  title     = {{CTR}-Guided Generative Query Suggestion in Conversational Search},
  booktitle = {Proceedings of the Conference on Empirical Methods in Natural Language Processing: Industry Track (EMNLP Industry)},
  year      = {2025},
  pages     = {2624--2634},
  doi       = {10.18653/v1/2025.emnlp-industry.178},
  publisher = {Association for Computational Linguistics}
}

@article{gqr2025,
  author  = {Min, Erxue and Huang, Hsiu-Yuan and Yang, Min and Yang, Xihong and Jia, Xin and Wu, Yunfang and Cai, Hengyi and Wang, Shuaiqiang and Yin, Dawei},
  title   = {From Prompting to Alignment: A Generative Framework for Query Recommendation},
  journal = {arXiv preprint arXiv:2504.10208},
  year    = {2025}
}

@inproceedings{relist2026,
  author    = {Bi, Shuxian and Wang, Chenxu and Wang, Wenjie and Mou, Yueqi and Feng, Fuli and Tang, Biao and Yan, Peng},
  title     = {{ReList}: A Multi-objective Reasoning Framework for Diversified Listwise Query Recommendation},
  booktitle = {Proceedings of the Annual Meeting of the Association for Computational Linguistics: Industry Track (ACL Industry)},
  year      = {2026},
  pages     = {1392--1405},
  doi       = {10.18653/v1/2026.acl-industry.97},
  publisher = {Association for Computational Linguistics}
}

@article{onepred2026,
  author  = {Chen, Jiangwang and Zhang, Bowen and Song, Zixin and Kang, Jiazheng and Yang, Xiao and Zhu, Da and Jiang, Guanjun},
  title   = {{OnePred}: Next-Query Prediction via Recursive Intent Memory in Multi-Turn Conversations},
  journal = {arXiv preprint arXiv:2605.23668},
  year    = {2026}
}

@inproceedings{kuaishou2026,
  author    = {Tian, Mingkai and {Xuye} and Meng, Long and Chen, Liwei and Qin, Zhiheng and Wang, Yi},
  title     = {From Short Video to Clickable Search: {RLVR}-Enabled Listwise Query Suggestion with Retrieval-Augmented Context},
  booktitle = {Proceedings of the Annual Meeting of the Association for Computational Linguistics: Industry Track (ACL Industry)},
  year      = {2026},
  pages     = {552--562},
  doi       = {10.18653/v1/2026.acl-industry.38},
  publisher = {Association for Computational Linguistics}
}

@inproceedings{icebreaker2026,
  author    = {Zheng, Hongwei and Wu, Weiqi and Wang, Zhengjia and Jiang, Guanyu and Li, Haoming and Wu, Tianyu and Zhu, Yongchun and Chen, Jingwu and Zhang, Feng},
  title     = {{IceBreaker} for Conversational Agents: Breaking the First-Message Barrier with Personalized Starters},
  booktitle = {Proceedings of the Annual Meeting of the Association for Computational Linguistics: Industry Track (ACL Industry)},
  year      = {2026},
  pages     = {230--241},
  doi       = {10.18653/v1/2026.acl-industry.16},
  publisher = {Association for Computational Linguistics}
}

@inproceedings{rosset2020leading,
  author    = {Rosset, Corby and Xiong, Chenyan and Song, Xia and Campos, Daniel and Craswell, Nick and Tiwary, Saurabh and Bennett, Paul},
  title     = {Leading Conversational Search by Suggesting Useful Questions},
  booktitle = {Proceedings of the ACM Web Conference (WWW)},
  year      = {2020},
  pages     = {1160--1170},
  doi       = {10.1145/3366423.3380193},
  publisher = {ACM},
  address   = {New York, NY, USA}
}

@inproceedings{wang2024mmqs,
  author    = {Wang, Zheng and Gan, Bingzheng and Shi, Wei},
  title     = {Multimodal Query Suggestion with Multi-Agent Reinforcement Learning from Human Feedback},
  booktitle = {Proceedings of the ACM Web Conference (WWW)},
  year      = {2024},
  pages     = {1374--1385},
  doi       = {10.1145/3589334.3645365},
  publisher = {ACM},
  address   = {New York, NY, USA}
}

@inproceedings{cao2008context,
  author    = {Cao, Huanhuan and Jiang, Daxin and Pei, Jian and He, Qi and Liao, Zhen and Chen, Enhong and Li, Hang},
  title     = {Context-Aware Query Suggestion by Mining Click-Through and Session Data},
  booktitle = {Proceedings of the ACM SIGKDD International Conference on Knowledge Discovery and Data Mining (KDD)},
  year      = {2008},
  pages     = {875--883},
  doi       = {10.1145/1401890.1401995},
  publisher = {ACM},
  address   = {New York, NY, USA}
}

@inproceedings{boldi2008queryflow,
  author    = {Boldi, Paolo and Bonchi, Francesco and Castillo, Carlos and Donato, Debora and Gionis, Aristides and Vigna, Sebastiano},
  title     = {The Query-Flow Graph: Model and Applications},
  booktitle = {Proceedings of the ACM International Conference on Information and Knowledge Management (CIKM)},
  year      = {2008},
  pages     = {609--618},
  doi       = {10.1145/1458082.1458163},
  publisher = {ACM},
  address   = {New York, NY, USA}
}

@inproceedings{sordoni2015hred,
  author    = {Sordoni, Alessandro and Bengio, Yoshua and Vahabi, Hossein and Lioma, Christina and Simonsen, Jakob Grue and Nie, Jian-Yun},
  title     = {A Hierarchical Recurrent Encoder-Decoder for Generative Context-Aware Query Suggestion},
  booktitle = {Proceedings of the ACM International Conference on Information and Knowledge Management (CIKM)},
  year      = {2015},
  pages     = {553--562},
  doi       = {10.1145/2806416.2806493},
  publisher = {ACM},
  address   = {New York, NY, USA}
}

@inproceedings{brooks2023instructpix2pix,
  author    = {Brooks, Tim and Holynski, Aleksander and Efros, Alexei A.},
  title     = {{InstructPix2Pix}: Learning to Follow Image Editing Instructions},
  booktitle = {Proceedings of the IEEE/CVF Conference on Computer Vision and Pattern Recognition (CVPR)},
  year      = {2023},
  pages     = {18392--18402},
  publisher = {IEEE}
}

@article{shen2024creativity,
  author  = {Shen, Tiancheng and Liew, Jun Hao and Mai, Long and Qi, Lu and Feng, Jiashi and Jia, Jiaya},
  title   = {Empowering Visual Creativity: A Vision-Language Assistant to Image Editing Recommendations},
  journal = {arXiv preprint arXiv:2406.00121},
  year    = {2024}
}

@inproceedings{zeng2026mira,
  author    = {Zeng, Ziyun and Hua, Hang and Luo, Jiebo},
  title     = {{MIRA}: Multimodal Iterative Reasoning Agent for Image Editing},
  booktitle = {Proceedings of the IEEE/CVF Conference on Computer Vision and Pattern Recognition (CVPR) Findings},
  year      = {2026},
  pages     = {9563--9573},
  publisher = {IEEE}
}

@inproceedings{joachims2002optimizing,
  author    = {Joachims, Thorsten},
  title     = {Optimizing Search Engines using Clickthrough Data},
  booktitle = {Proceedings of the ACM SIGKDD International Conference on Knowledge Discovery and Data Mining (KDD)},
  year      = {2002},
  pages     = {133--142},
  doi       = {10.1145/775047.775067},
  publisher = {ACM},
  address   = {New York, NY, USA}
}

@inproceedings{joachims2005accurately,
  author    = {Joachims, Thorsten and Granka, Laura and Pan, Bing and Hembrooke, Helene and Gay, Geri},
  title     = {Accurately Interpreting Clickthrough Data as Implicit Feedback},
  booktitle = {Proceedings of the International ACM SIGIR Conference on Research and Development in Information Retrieval (SIGIR)},
  year      = {2005},
  pages     = {154--161},
  doi       = {10.1145/1076034.1076063},
  publisher = {ACM},
  address   = {New York, NY, USA}
}

@inproceedings{craswell2008experimental,
  author    = {Craswell, Nick and Zoeter, Onno and Taylor, Michael and Ramsey, Bill},
  title     = {An Experimental Comparison of Click Position-Bias Models},
  booktitle = {Proceedings of the ACM International Conference on Web Search and Data Mining (WSDM)},
  year      = {2008},
  pages     = {87--94},
  doi       = {10.1145/1341531.1341545},
  publisher = {ACM},
  address   = {New York, NY, USA}
}

@inproceedings{joachims2017unbiased,
  author    = {Joachims, Thorsten and Swaminathan, Adith and Schnabel, Tobias},
  title     = {Unbiased Learning-to-Rank with Biased Feedback},
  booktitle = {Proceedings of the ACM International Conference on Web Search and Data Mining (WSDM)},
  year      = {2017},
  pages     = {781--789},
  doi       = {10.1145/3018661.3018699},
  publisher = {ACM},
  address   = {New York, NY, USA}
}

@inproceedings{wang2018position,
  author    = {Wang, Xuanhui and Golbandi, Nadav and Bendersky, Michael and Metzler, Donald and Najork, Marc},
  title     = {Position Bias Estimation for Unbiased Learning to Rank in Personal Search},
  booktitle = {Proceedings of the ACM International Conference on Web Search and Data Mining (WSDM)},
  year      = {2018},
  pages     = {610--618},
  doi       = {10.1145/3159652.3159732},
  publisher = {ACM},
  address   = {New York, NY, USA}
}

@inproceedings{ai2018unbiased,
  author    = {Ai, Qingyao and Bi, Keping and Luo, Cheng and Guo, Jiafeng and Croft, W. Bruce},
  title     = {Unbiased Learning to Rank with Unbiased Propensity Estimation},
  booktitle = {Proceedings of the International ACM SIGIR Conference on Research and Development in Information Retrieval (SIGIR)},
  year      = {2018},
  pages     = {385--394},
  doi       = {10.1145/3209978.3209986},
  publisher = {ACM},
  address   = {New York, NY, USA}
}

@article{bradley1952rank,
  author  = {Bradley, Ralph Allan and Terry, Milton E.},
  title   = {Rank Analysis of Incomplete Block Designs: I. The Method of Paired Comparisons},
  journal = {Biometrika},
  volume  = {39},
  number  = {3/4},
  pages   = {324--345},
  year    = {1952}
}

@inproceedings{goyal2017vqa,
  author    = {Goyal, Yash and Khot, Tejas and Summers-Stay, Douglas and Batra, Dhruv and Parikh, Devi},
  title     = {Making the {V} in {VQA} Matter: Elevating the Role of Image Understanding in Visual Question Answering},
  booktitle = {Proceedings of the IEEE Conference on Computer Vision and Pattern Recognition (CVPR)},
  year      = {2017},
  pages     = {6325--6334},
  publisher = {IEEE}
}

@inproceedings{agrawal2018vqacp,
  author    = {Agrawal, Aishwarya and Batra, Dhruv and Parikh, Devi and Kembhavi, Aniruddha},
  title     = {Don't Just Assume; Look and Answer: Overcoming Priors for Visual Question Answering},
  booktitle = {Proceedings of the IEEE/CVF Conference on Computer Vision and Pattern Recognition (CVPR)},
  year      = {2018},
  pages     = {4971--4980},
  publisher = {IEEE}
}

@inproceedings{rohrbach2018object,
  author    = {Rohrbach, Anna and Hendricks, Lisa Anne and Burns, Kaylee and Darrell, Trevor and Saenko, Kate},
  title     = {Object Hallucination in Image Captioning},
  booktitle = {Proceedings of the Conference on Empirical Methods in Natural Language Processing (EMNLP)},
  year      = {2018},
  pages     = {4035--4045},
  doi       = {10.18653/v1/D18-1437},
  publisher = {Association for Computational Linguistics}
}

@inproceedings{pope2023,
  author    = {Li, Yifan and Du, Yifan and Zhou, Kun and Wang, Jinpeng and Zhao, Wayne Xin and Wen, Ji-Rong},
  title     = {Evaluating Object Hallucination in Large Vision-Language Models},
  booktitle = {Proceedings of the Conference on Empirical Methods in Natural Language Processing (EMNLP)},
  year      = {2023},
  pages     = {292--305},
  doi       = {10.18653/v1/2023.emnlp-main.20},
  publisher = {Association for Computational Linguistics}
}

@inproceedings{vcd2024,
  author    = {Leng, Sicong and Zhang, Hang and Chen, Guanzheng and Li, Xin and Lu, Shijian and Miao, Chunyan and Bing, Lidong},
  title     = {Mitigating Object Hallucinations in Large Vision-Language Models through Visual Contrastive Decoding},
  booktitle = {Proceedings of the IEEE/CVF Conference on Computer Vision and Pattern Recognition (CVPR)},
  year      = {2024},
  pages     = {13872--13882},
  publisher = {IEEE}
}

@inproceedings{hallusionbench2024,
  author    = {Guan, Tianrui and Liu, Fuxiao and Wu, Xiyang and Xian, Ruiqi and Li, Zongxia and Liu, Xiaoyu and Wang, Xijun and Chen, Lichang and Huang, Furong and Yacoob, Yaser and Manocha, Dinesh and Zhou, Tianyi},
  title     = {{HallusionBench}: An Advanced Diagnostic Suite for Entangled Language Hallucination and Visual Illusion in Large Vision-Language Models},
  booktitle = {Proceedings of the IEEE/CVF Conference on Computer Vision and Pattern Recognition (CVPR)},
  year      = {2024},
  pages     = {14375--14385},
  publisher = {IEEE}
}

@inproceedings{sesame2024,
  author    = {Wu, Tsung-Han and Biamby, Giscard and Chan, David M. and Dunlap, Lisa and Gupta, Ritwik and Wang, Xudong and Gonzalez, Joseph E. and Darrell, Trevor},
  title     = {See, Say, and Segment: Teaching {LMMs} to Overcome False Premises},
  booktitle = {Proceedings of the IEEE/CVF Conference on Computer Vision and Pattern Recognition (CVPR)},
  year      = {2024},
  pages     = {13459--13469},
  publisher = {IEEE}
}

@inproceedings{rlhfv2024,
  author    = {Yu, Tianyu and Yao, Yuan and Zhang, Haoye and He, Taiwen and Han, Yifeng and Cui, Ganqu and Hu, Jinyi and Liu, Zhiyuan and Zheng, Hai-Tao and Sun, Maosong and Chua, Tat-Seng},
  title     = {{RLHF-V}: Towards Trustworthy {MLLMs} via Behavior Alignment from Fine-grained Correctional Human Feedback},
  booktitle = {Proceedings of the IEEE/CVF Conference on Computer Vision and Pattern Recognition (CVPR)},
  year      = {2024},
  pages     = {13807--13816},
  publisher = {IEEE}
}

@inproceedings{rlaifv2024,
  author    = {Yu, Tianyu and Zhang, Haoye and Li, Qiming and Xu, Qixin and Yao, Yuan and Chen, Da and Lu, Xiaoman and Cui, Ganqu and Dang, Yunkai and He, Taiwen and Feng, Xiaocheng and Song, Jun and Zheng, Bo and Liu, Zhiyuan and Chua, Tat-Seng and Sun, Maosong},
  title     = {{RLAIF-V}: Open-Source {AI} Feedback Leads to Super {GPT-4V} Trustworthiness},
  booktitle = {Proceedings of the IEEE/CVF Conference on Computer Vision and Pattern Recognition (CVPR)},
  year      = {2025},
  pages     = {19985--19995},
  publisher = {IEEE}
}

@inproceedings{hadpo2023,
  author    = {Zhao, Zhiyuan and Wang, Bin and Ouyang, Linke and Dong, Xiaoyi and Wang, Jiaqi and He, Conghui},
  title     = {Beyond Multimodal Hallucinations: Enhancing {LVLMs} through Hallucination-Aware Direct Preference Optimization},
  booktitle = {Proceedings of the IEEE International Conference on Multimedia and Expo (ICME)},
  year      = {2025},
  pages     = {1--6},
  doi       = {10.1109/ICME59968.2025.11209377},
  publisher = {IEEE}
}

@inproceedings{ouyang2022instructgpt,
  author    = {Ouyang, Long and Wu, Jeffrey and Jiang, Xu and Almeida, Diogo and Wainwright, Carroll L. and Mishkin, Pamela and Zhang, Chong and Agarwal, Sandhini and Slama, Katarina and Ray, Alex and Schulman, John and Hilton, Jacob and Kelton, Fraser and Miller, Luke and Simens, Maddie and Askell, Amanda and Welinder, Peter and Christiano, Paul F. and Leike, Jan and Lowe, Ryan},
  title     = {Training Language Models to Follow Instructions with Human Feedback},
  booktitle = {Advances in Neural Information Processing Systems (NeurIPS)},
  year      = {2022},
  pages     = {27730--27744}
}

@article{grpo2024,
  author  = {Shao, Zhihong and Wang, Peiyi and Zhu, Qihao and Xu, Runxin and Song, Junxiao and Bi, Xiao and Zhang, Haowei and Zhang, Mingchuan and Li, Y.K. and Wu, Y. and Guo, Daya},
  title   = {{DeepSeekMath}: Pushing the Limits of Mathematical Reasoning in Open Language Models},
  journal = {arXiv preprint arXiv:2402.03300},
  year    = {2024}
}

@inproceedings{odin2024,
  author    = {Chen, Lichang and Zhu, Chen and Soselia, Davit and Chen, Jiuhai and Zhou, Tianyi and Goldstein, Tom and Huang, Heng and Shoeybi, Mohammad and Catanzaro, Bryan},
  title     = {{ODIN}: Disentangled Reward Mitigates Hacking in {RLHF}},
  booktitle = {Proceedings of the International Conference on Machine Learning (ICML)},
  year      = {2024},
  pages     = {7935--7952}
}

@article{gdpo2026,
  author  = {Liu, Shih-Yang and Dong, Xin and Lu, Ximing and Diao, Shizhe and Belcak, Peter and Liu, Mingjie and Chen, Min-Hung and Yin, Hongxu and Wang, Yu-Chiang Frank and Cheng, Kwang-Ting and Choi, Yejin and Kautz, Jan and Molchanov, Pavlo},
  title   = {{GDPO}: Group Reward-Decoupled Normalization Policy Optimization for Multi-Reward {RL} Optimization},
  journal = {arXiv preprint arXiv:2601.05242},
  year    = {2026}
}

@article{dvao2026,
  author  = {Jiang, Guochao and Song, Jingyi and Quan, Guofeng and Hao, Chuzhan and Liu, Guohua and Zhang, Yuewei},
  title   = {{DVAO}: Dynamic Variance-Adaptive Advantage Optimization for Multi-Reward Reinforcement Learning},
  journal = {arXiv preprint arXiv:2605.25604},
  year    = {2026}
}

@article{mogrpo2025,
  author  = {Ichihara, Yuki and Jinnai, Yuu and Morimura, Tetsuro and Sakamoto, Mitsuki and Mitsuhashi, Ryota and Uchibe, Eiji},
  title   = {{MO-GRPO}: Mitigating Reward Hacking of Group Relative Policy Optimization on Multi-Objective Problems},
  journal = {arXiv preprint arXiv:2509.22047},
  year    = {2025},
  note    = {Accepted by TACL}
}

@article{dynweight2026,
  author  = {Lu, Yining and Wang, Zilong and Li, Shiyang and Liu, Xin and Yu, Changlong and Yin, Qingyu and Shi, Zhan and Zhang, Zixuan and Jiang, Meng},
  title   = {Learning to Optimize Multi-Objective Alignment Through Dynamic Reward Weighting},
  journal = {Transactions of the Association for Computational Linguistics},
  year    = {2026},
  note    = {arXiv:2509.11452}
}

@article{ctwa2026,
  author  = {Lu, Yining and Jiang, Meng},
  title   = {Uncovering Cross-Objective Interference in Multi-Objective Alignment},
  journal = {arXiv preprint arXiv:2602.06869},
  year    = {2026}
}

@inproceedings{dqo2026,
  author    = {Chen, Yilei and Chakraborty, Souradip and Wolf, Lorenz and Paschalidis, Ioannis and Pacchiano, Aldo},
  title     = {Post-Training Large Language Models for Diverse High-Quality Responses},
  booktitle = {Proceedings of the International Conference on Learning Representations (ICLR)},
  year      = {2026},
  note      = {arXiv:2509.04784}
}

@inproceedings{hu2022lora,
  author    = {Hu, Edward J. and Shen, Yelong and Wallis, Phillip and Allen-Zhu, Zeyuan and Li, Yuanzhi and Wang, Shean and Wang, Lu and Chen, Weizhu},
  title     = {{LoRA}: Low-Rank Adaptation of Large Language Models},
  booktitle = {Proceedings of the International Conference on Learning Representations (ICLR)},
  year      = {2022}
}

@article{qwen3vl2025,
  author  = {Bai, Shuai and others},
  title   = {{Qwen3-VL} Technical Report},
  journal = {arXiv preprint arXiv:2511.21631},
  year    = {2025}
}

@article{qwen3embedding2025,
  author  = {Zhang, Yanzhao and Li, Mingxin and Long, Dingkun and Zhang, Xin and Lin, Huan and Yang, Baosong and Xie, Pengjun and Yang, An and Liu, Dayiheng and Lin, Junyang and Huang, Fei and Zhou, Jingren},
  title   = {{Qwen3 Embedding}: Advancing Text Embedding and Reranking Through Foundation Models},
  journal = {arXiv preprint arXiv:2506.05176},
  year    = {2025}
}

@misc{gemini31pro2026,
  author       = {{Google DeepMind}},
  title        = {{Gemini 3.1 Pro} Model Card},
  year         = {2026},
  howpublished = {\url{https://deepmind.google/models/model-cards/gemini-3-1-pro/}},
  note         = {Preview release}
}

@misc{gemini3flash2025,
  author       = {{Google DeepMind}},
  title        = {{Gemini 3 Flash} Model Card},
  year         = {2025},
  howpublished = {\url{https://deepmind.google/models/model-cards/gemini-3-flash/}},
  note         = {Updated 17 December 2025}
}

\clearpage
\appendix

\section{Reproducibility and Data Protocol}
\label{app:data}

\subsection{Splitting and Leakage Controls}

The preference pipeline keeps at most one eligible impression for each non-empty session ID before pairing. It then constructs all position-aware pairs and splits them by request ID with seed 42. The resulting train and validation sets share no request ID. Among the 128{,}750 paired impressions, 22 have an empty session ID; these can be separated only by request ID. No usable user ID is present, so we do not claim a user-level split. SFT and RL follow their own request-level splits, and the 500 policy-evaluation sessions come from a disjoint time window and are excluded from training. All data are anonymized under the applicable internal use and retention policies.

\subsection{Generation Tracks}
\label{app:gentracks}

The general track covers all retained contexts. The priority track gives more specific directions for common high-traffic cases. The continuation track supplies additional candidates only when validation leaves fewer than six suggestions. All three tracks remain subject to the same intent-label, validator, and product-rule checks.

\subsection{Preference-Data Funnel}

Table~\ref{tab:datafunnel} summarizes the filtering pipeline. The source contains clicked requests already grouped by request ID and restricted to three displayed suggestions with a non-empty click. The production display layer randomly sampled these suggestions without replacement from the generator slate. The file therefore has no unclicked-impression denominator.

\begin{table*}[t]
\caption{Preference-data funnel. Counts in the first three rows are impression-level unless marked as suggestion-level.}
\label{tab:datafunnel}
\centering
\footnotesize
\setlength{\tabcolsep}{4pt}
\begin{tabular}{@{}p{0.13\textwidth}p{0.55\textwidth}r@{}}
\toprule
Stage & Filtering operation & Output\\
\midrule
Source & Three displayed suggestions and a non-empty click & 663{,}926 requests\\
Record filtering & Remove click not in display (1{,}861), invalid intent (45{,}022), invalid clicked type/sub-tag (120{,}344), fewer than two usable continuation suggestions (15{,}344), and repeated sessions (249{,}062) & 232{,}293 impressions\\
Image transfer & Retain records with a successfully transferred image & 232{,}281 impressions\\
Pair eligibility & Remove unusable query (944), query-copy click (66), missing final intent (30), and no usable suggestion above the click (102{,}491) & 128{,}750 impressions\\
Suggestion pruning & Drop generic refresh-like (39{,}829), non-continuation (45{,}420), and below-click options (204{,}173); these are suggestion-level counts & 173{,}071 pairs\\
Request split & Group by request ID & 164{,}401 train / 8{,}670 validation pairs\\
\bottomrule
\end{tabular}
\end{table*}

Of the 128{,}750 eligible impressions, 84{,}429 produce one pair and 44{,}321 produce two. The request-grouped split contains 122{,}313 training requests and 6{,}437 validation requests.

\subsection{Denominators and Statistics}

Policy generation draws four samples for each of the 500 evaluation sessions. Judge cost then sets the denominator of each automated audit. The cross-checkpoint grounding audit of Appendix~\ref{app:groundaudit} scores the first two samples and keeps the 970 image--sample units on which every checkpoint returned a valid external judgment. These units contain 5{,}811 suggestions for SFT, 5{,}725 for Stage 2, and 5{,}716 for the full framework. Because the audited sample indices are fixed in advance, they are not selected by difficulty. Redundancy and cross-image reuse use all 2{,}000 units.

Among the SFT suggestions, 172 are flagged across 143 slates. Visual inconsistencies are therefore sparse and local, at roughly one flagged suggestion per affected slate. This pattern supports the suggestion-level aggregation in Equation~\ref{eq:ground}.

The visual-consistency calibration set holds 488 visual inconsistencies, of which 112 are missing sources and 376 are already-satisfied targets, together with 491 expert-labeled excellent suggestions. The multi-turn query audit covers 100{,}000 adjacent-turn pairs, 60{,}837 of them after keeping one pair per session. Separately, the three position-aware win rates from the click-preference data rest on 7{,}812, 19{,}783, and 1{,}030 pairs, respectively.

\subsection{Multi-Turn Query Audit of Image Dependence}

The audit reported in Section~\ref{sec:intro} samples 100{,}000 adjacent user-turn pairs from image-generation conversations. The later turn is the observed follow-up editing query after the user sees the image produced for the previous query. An image-dependent query introduces concrete content absent from the previous query and needs the latest image for grounding. A text-dependent query is supported by the previous query or uses a generic edit template.

A deterministic lexicon and pattern classifier assigns 80.1\% and 19.9\% of queries to these two classes. At least 23.9 percentage points of all follow-up editing queries act on a specific source object in the image, whereas 3.8 percentage points are generic edits applicable across images. The former is a conservative lower bound because ambiguous references are omitted.

The results remain stable after keeping one adjacent-turn pair per session and on a separate 10K sample. The current query is rewritten from the multi-turn dialogue to incorporate relevant prior context; neither the teacher nor the deployed policy receives the raw earlier turns.

In a separate comparison against the fixed template pool, source-referential, target-novel, and generic suggestions win 56.7\%, 55.1\%, and 55.2\% of position-aware click pairs, respectively. These win rates are not part of the 100K image-dependence classification.

\subsection{External Grounding Audit}
\label{app:groundaudit}

The grounding column of Tables~\ref{tab:main} and~\ref{tab:rewardablation} is produced outside the training loop. Every arm, including PE, is scored by Gemini 3.1 Pro \cite{gemini31pro2026} under the same audit protocol; the PE rate reported in Table~\ref{tab:main} is therefore directly comparable with the learned policies. Gemini comes from a different model family and provider than the Qwen3-VL-30B-A3B verifier that supplies the reward, carries separate parameters and evaluation prompts, and takes no part in policy training or checkpoint selection. Arms are scored in one batch on identical units, with arm identity withheld from the judge.

Aggregate agreement alone cannot validate a rare-event audit: a judge that never flags an inconsistency would still agree with most decisions. We therefore sample the two judge strata separately. Experts adjudicate all 435 flagged suggestions and a stratified random sample of 841 cleared suggestions: 280 from SFT, 280 from Stage 2, and 281 from the full framework. Reviewers see the latest image, query, intent, and suggestion under the rubric of Appendix~\ref{app:human}, but neither the policy arm nor the judge decision.

Flagged suggestions are reviewed exhaustively, whereas cleared suggestions are sampled. We therefore estimate recall with inverse-probability weights instead of reading it directly from the reviewed counts; one reviewed cleared suggestion represents about twenty suggestions in its stratum. On the visual-inconsistency class, the judge reaches 90.7\% recall and 89.0\% precision, with 96.1\% agreement over the 1{,}276 reviewed suggestions.

Balanced agreement, the mean agreement on flagged and cleared suggestions, is 94.3\%, 94.4\%, and 94.1\% across the three arms. Judge-call failures are 2.0\%, 1.7\%, and 1.8\%, so neither error nor attrition follows the policy arm. Applying the same weights to the expert labels gives population visual-inconsistency rates of 3.0\%, 3.7\%, and 0.8\%. The expert-corrected gap between Stage 2 and the full framework is thus 2.9 points, close to the judge-reported 2.8 points.

The training verifier still scores policy output during checkpoint selection, but those numbers are operational and are not reported as results.

\subsection{Training Configuration}
\label{app:training}

\par\textbf{SFT.}
We train Qwen3-VL-8B for five epochs at learning rate $10^{-5}$ with LoRA rank 4 and $\alpha=16$, a frozen visual encoder \cite{hu2022lora}, sequence length 8192, per-device batch size 1, gradient accumulation 2, and 5\% warmup.

\par\textbf{Reward Model.}
The 8B click RM trains for one epoch at learning rate $5\times10^{-5}$ with LoRA rank 16 and $\alpha=64$. We target all linear modules, freeze the visual encoder, train the multimodal aligner, use sequence length 4096, and set the effective batch size to 32.

\par\textbf{Reinforcement Learning.}
RL samples $G=8$ rollout slates per context with a global batch of 128, GRPO minibatches of 32, one epoch at learning rate $10^{-6}$, and 5\% warmup. The KL and entropy coefficients are 0.15 and 0.001, the GRPO clip half-width is $\epsilon_{\mathrm{clip}}=0.2$, and the reward-normalization floor is $\epsilon_\sigma=10^{-6}$.

Every arm shares actor initialization, reward model, data order, batch size, learning rate, rollout count, verifier, and evaluation requests. Generation for evaluation uses the same decoder for all arms: temperature 0.7, nucleus sampling at 0.9, at most 512 new tokens, a visual token budget of 1{,}254{,}400 pixels, and seed 42. Seeds are not reproducible across runs on our serving stack, so all arms in a comparison are regenerated in one batch and scored together.

\section{Source--Target Representation and Verifier}
\label{app:verifier}

The VLM judge follows an image-first decision procedure rather than a single end-to-end validity prompt. It first records the current visual state, then extracts required sources and a target state from each candidate text, and finally checks source existence and target satisfaction separately. The two resulting flags determine the grounding reward. We describe this decision contract and its structured outputs below, but omit verbatim production prompts, product-specific lexicons, and exact post-processing triggers.

\subsection{Deployed Output Schema}

One verifier call scores a whole slate. It first emits the candidate-independent image inventory, then one record per candidate:
\begin{verbatim}
{
"image_inventory": {
  "main_subjects": "...", "visible_text": [],
  "watermark": "...",     "background": "...",
  "notable_objects": [],
  "scene_state": {"background": "...",
    "style": "...", "layout": "...",
    "person": "..."}},
"per_item": [{
  "sources":  ["black hoodie"],
  "targets":  ["white shirt"],
  "exists":   [1],
  "evidence": ["left figure, black hoodie"],
  "target_state":  "shirt = white",
  "current_state": "hoodie = black",
  "already_satisfied": 0,
  "plausible": 1.0, "effective": 1.0,
  "content_units": 1, "reason": "..."}]
}
\end{verbatim}
Sources are checked for presence in the image; targets are not required to be present. Instead, visually decidable target states are compared with the current state to detect an already-satisfied edit. Every source needs an existence bit and a short pointing-evidence phrase. The missing-source flag comes from the existence bits, and the already-satisfied flag comes from the state comparison; the model's free-form summary is ignored. Global edits may have no localized source. When target satisfaction is subjective or not visually decidable, the state fields stay empty and the reward fails open.

\subsection{Constructed Visual Inconsistency Cases}

Expert labels alone yield too few visual inconsistencies to study the two types separately. We therefore augment the calibration set with production editing chains whose labels follow from edit semantics, without using a model to label them. Because an editing model can fail or only partially execute an instruction, we first discard chains whose output does not visibly contain the requested change and manually confirm all retained cases.

For a retained chain, the output image visibly contains the requested change. Restating the original instruction against that output is therefore already satisfied by construction.

An object introduced by the original instruction is absent from the input image. We write a new instruction that removes or modifies this object and evaluate it against the input, where the required source is guaranteed to be missing.

Only instructions with a discrete target state qualify, because repeating a relative or subjective instruction does not form a clean no-op. Missing-source construction also requires the input image, which is the scarcer side of the logs; this subset is therefore smaller. These constructed cases complement rather than replace evaluation on naturally occurring policy outputs.

\subsection{Transfer of the Calibration Rates}

The calibration set is not sampled from production traffic. Its visual inconsistencies are gathered or constructed as such, and its excellent items come from expert worksheets. Their ratio therefore says nothing about online prevalence.

Recall and false rejection are conditional on the true label and remain meaningful under this shift, so we report both. Precision and other prevalence-dependent quantities do not transfer. The visual-inconsistency rate on real policy output instead comes from the external audit in Table~\ref{tab:main}, which uses a different judge (Appendix~\ref{app:groundaudit}).

Failed or unparseable calls are retried until every candidate has a record from both verifiers. The two systems are therefore compared on identical items.

The comparison evaluates the source--target verifier end to end and does not attribute the gain among inventory ordering, source and target decomposition, and post-parsing guards, since all three change together between the two prompts.

\subsection{Verifier Decision Procedure}

The controlled comparison in Section~\ref{subsec:verifierstudy} compares the source--target procedure with a single-pass procedure. The ordered steps below define the source--target decision contract used by the complete production configuration.

A candidate is an instruction to be applied to the image, not a description or a question. Asking for a state that differs from the current image is normal editing and must not be penalized. Neither the user's request nor confident candidate wording proves that an object is present. Only visual evidence recorded by the verifier counts.

\par\textbf{Step 1: Image Observation.}
Before reading any candidate, the verifier records the main subjects, visible text quoted verbatim, watermark, background, notable objects, and a scene state covering background, style, layout, and person. It uses a conservative rule: content not visibly supported is absent.

\par\textbf{Step 2: Source--Target Decomposition.}
For each candidate, the verifier asks whether each noun phrase must already occur in the image or is introduced by the edit. The former is a source; the latter is a target. Thus, an added object is a target, whereas an object removed or modified is a source. Targets are not required to exist beforehand. Sources must be grounded in candidate text, and attributes or states cannot serve as standalone source objects. Product-specific lexical disambiguation handles implicit, global, and interface references without changing these rules.

\par\textbf{Step 3: Source and Target Validation.}
The verifier revisits the image for each source rather than trusting the coarse inventory. A present source requires a short location and appearance phrase. An instance of the source category is sufficient even if its attributes differ; a category with no visible instance is missing. For visually decidable targets, the verifier writes the desired and current states side by side and compares them.

\par\textbf{Step 4: Visual Inconsistency Detection.}
The verifier derives the final inconsistency flags from the source-existence and target-state fields. One-directional post-processing may clear known parser-induced false alarms but cannot create a new inconsistency.

\par\textbf{Single-Pass Baseline.}
Our prior production rubric uses the same backbone, image, and candidates, but runs in one pass without an inventory or source--target split. It scores whether an edit is reasonable and executable and whether it avoids contradicting the image while producing a real change, each in $\{0,0.5,1\}$; either score at zero indicates a visual inconsistency. The baseline receives the same instruction not to penalize a requested change merely for differing from the current image, and the same already-satisfied examples. The controlled variable is therefore the ordered decomposition.

\subsection{Judged Dimensions and Their Roles}

\begin{table}[h]
\caption{Per-candidate dimensions emitted by one verifier call.}
\label{tab:dimensions}
\reportfloatwidth{1.00}
\small
\begin{tabular}{@{}l
  >{\raggedright\arraybackslash}p{0.23\columnwidth}
  >{\raggedright\arraybackslash}p{0.3\columnwidth}@{}}
\toprule
Dimension & Judged from & Used for\\
\midrule
missing source & source existence bits with evidence & RL grounding $\widehat g$\\
already satisfied & target vs.\ current state & RL grounding $\widehat g$\\
plausibility & fit to image and scene & monitoring only\\
effective change & non-trivial visible change & monitoring only\\
content density & global appearance vs.\ object/region edit & length budget class\\
evidence, states & pointing phrases & audits and error analysis\\
\bottomrule
\end{tabular}
\end{table}

Table~\ref{tab:dimensions} lists the per-candidate dimensions. Only the two inconsistency flags enter the grounding reward. Plausibility and effective-change scores are emitted for monitoring and saturate near the maximum on normal slates, which is expected for an audit signal. The content-density class only selects the length budget in Equation~\ref{eq:length}. Global appearance edits, such as filters, style transfer, color grading, upscaling, and cropping, use the shorter budget. An edit that names, adds, removes, or changes an object or region uses the longer budget when that object is grounded in the image. Otherwise it falls back to the shorter budget, so inventing an object cannot earn extra length.

\subsection{One-Directional Guard Design}

Post-parsing guards address a small set of known false-positive families, such as source--target role confusion, implicit or global references, and targets without a discrete visually decidable state. The exact production triggers are product-specific and are not reproduced. The important design invariant is one-directionality: a guard may clear a suspected inconsistency but may never create one. We tested a bidirectional alternative that also allowed deterministic rules to add inconsistency flags. Although it raised recall on targeted probes, it tripled false rejection on excellent suggestions. We therefore retain the clearing-only design.

\subsection{Operation Semantics}

\begin{table}[h]
\caption{Source and target semantics by edit type.}
\label{tab:opsemantics}
\reportfloatwidth{1.00}
\small
\begin{tabular}{@{}l
  >{\raggedright\arraybackslash}p{0.27\columnwidth}
  >{\raggedright\arraybackslash}p{0.43\columnwidth}@{}}
\toprule
Operation & Required source & Target check\\
\midrule
Add & anchor/context entity & requested addition not satisfied\\
Remove & object to remove & object not already absent\\
Modify & object and current state & new state not already satisfied\\
Relocate & object and reference region & requested relation not satisfied\\
Global style & image & style change checkable or uncertain\\
\bottomrule
\end{tabular}
\end{table}

\section{Optimization Details}
\label{app:optimization}

\par\textbf{Candidate-to-Display Aggregation.} Under random sampling without replacement, the expected mean suggestion reward of the three displayed suggestions equals the mean over the candidate slate:
\begin{equation}
\mathbb E_D\!\left[\frac{1}{3}\sum_{j\in D}r_\phi(x,y^j)\right]
=\frac{1}{N}\sum_{j=1}^{N}r_\phi(x,y^j)
=\bar r_\phi(Y).
\label{eq:displayexpectation}
\end{equation}
This identity supports the additive suggestion-reward proxy; it does not assume that listwise CTR interactions decompose in the same way.

\par\textbf{Masked Grounding Normalization.} Let $m_i=\mathbf 1[r_{\mathrm{gate},i}=1]$ and $\mathcal V_x=\{i:m_i=1\}$ denote the valid rollouts in a group. For $i\in\mathcal V_x$, set $r_i^{\mathrm{grd}}=\widehat g_i$ and compute the grounding mean and standard deviation only over $\mathcal V_x$:
\begin{equation}
\widetilde A_i^{\mathrm{grd}}=
\begin{cases}
\dfrac{r_i^{\mathrm{grd}}-\mu_{\mathrm{grd},x}}
{\sigma_{\mathrm{grd},x}+\epsilon_\sigma},
& i\in\mathcal V_x,\ |\mathcal V_x|\ge 2,\\[6pt]
0, & \text{otherwise}.
\end{cases}
\label{eq:maskedground}
\end{equation}
Thus, gate-failed rollouts receive zero grounding advantage, and a group with fewer than two valid rollouts contributes no grounding update.

Sections~\ref{subsec:rewards} and~\ref{subsec:grounding} define all six rewards. For diversity, raw cosine similarities are linearly calibrated from $[a,b]=[0.3,0.9]$ and clipped to $[0,1]$:
\begin{align}
u_{jj'}&=\frac{\cos(e_j,e_{j'})-a}{b-a},\\
\widetilde s_{jj'}&=\operatorname{clip}(u_{jj'},0,1).
\end{align}
The gate accepts a JSON object with allowed intent labels and 5--7 parsed suggestions. This is the product-valid range, while prompting and supervised targets continue to favor six. Gate-failed rollouts do not invoke the visual verifier. They use gate, length, and diversity scores of zero and retain the RM score, but grounding is masked: it supplies neither a reward value nor an advantage. Grounding statistics are computed only over gate-passed rollouts in the group, as defined in Equation~\ref{eq:maskedground}.

The weights $\lambda_k$ for gate, preference, PPL, length, diversity, and grounding are initialized to $1.0$, $1.0$, $0.3$, $0.5$, $0.5$, and $0.5$, respectively. Gate, preference, PPL, and grounding remain fixed. For the two adaptive constraints in Equation~\ref{eq:textcontroller}, we use $\eta=\kappa=0.1$ and $\lambda_{\max}=1$. Length requires mean $r_{\mathrm{len}}\geq0.80$. Diversity limits the fraction of slates with $r_{\mathrm{div}}<0.2$ to $0.05$. A satisfied constraint therefore reduces its multiplier.

More explicitly,
\begin{equation}
\delta_{\mathrm{len}}=0.80-\widetilde m_{\mathrm{len}},
\qquad
\delta_{\mathrm{div}}=\widetilde m_{\mathrm{div}}-0.05,
\end{equation}
so either gap is positive exactly when its constraint is missed. The controller is updated once per training batch.

The length thresholds in Equation~\ref{eq:length} are display-budget constants of the deployed surface. The Qwen App recommendation rail renders Chinese text, so length is measured in characters: $L_1=11$ and $L_2=17$ for global appearance edits, and $L_1=13$ and $L_2=18$ for object- or region-level edits. The larger budget leaves room for a necessary object or region name.

The optimizer executes the following sequence for each batch:
\begin{enumerate}[leftmargin=*,topsep=2pt,itemsep=1pt]
\item sample $G$ slates from $\pi_\theta(\cdot\mid x)$;
\item score gate, preference, PPL, length, and diversity for every rollout, and score grounding only for gate-passed rollouts;
\item normalize the first five rewards separately within each rollout group; normalize grounding over gate-passed rollouts and assign zero grounding advantage to the rest;
\item update the length and diversity weights from their EMA constraint gaps;
\item take the weighted sum, apply masked batch standardization, and run the clipped GRPO update.
\end{enumerate}

\subsection{Length Preference of the Click Reward Model}
\label{app:lengthbias}

The length reward exists because the click reward model has a measurable preference for longer text. On ten meaning-preserving minimal pairs, each a suggestion paired with a padded rewrite that adds non-essential modifiers without changing the edit, the click reward model scores the padded version higher in all ten. Left unchecked, this bias transfers to the policy: preference-only optimization raises the mean suggestion length from 11.9 to 17.0 characters, again through non-essential modifiers rather than added content. The content-aware length reward of Equation~\ref{eq:length} counters this, while its larger budget for object- or region-level edits leaves room for necessary names.

\subsection{Diversity Signal and Aggregation}
\label{app:diversity}

The diversity reward in Equation~\ref{eq:div} uses embedding similarity rather than lexical overlap and max-pair aggregation rather than a mean. Table~\ref{tab:diversity} compares these choices on a semantic-redundancy probe. The probe is separate from the incremental chain in Table~\ref{tab:rewardablation}, but all five variants use one matched protocol.

Synonyms bypass lexical Jaccard, while mean embedding similarity dilutes one repeated pair among all $N(N-1)/2$ pairs. Both therefore leave redundancy at 28--30\%. Max-pair aggregation exposes the worst pair and lowers redundancy to 20.8\%. Adding the adaptive weight from Equation~\ref{eq:textcontroller} further lowers redundancy to 9.6\% and character-Jaccard near-duplicates to 0.4\%.

The matched-step results in Table~\ref{tab:rewardablation}, from 28.1\% to 19.1\%, show the same max-pair benefit inside the main training chain.

\begin{table}[t]
\caption{Diversity signal and aggregation on the semantic-redundancy probe.}
\label{tab:diversity}
\reportfloatwidth{0.50}
\small
\setlength{\tabcolsep}{6pt}
\begin{tabular}{@{}lccc@{}}
\toprule
Variant & Adapt.\ $\lambda$ & Redund. $\downarrow$ & Near-dup $\downarrow$ \\
\midrule
SFT & $\times$ & 14.8\% & 1.4\% \\
Jaccard & $\times$ & 28.0\% & 2.2\% \\
Embed., mean & $\times$ & 30.0\% & 2.2\% \\
Embed., max-pair & $\times$ & 20.8\% & 0.8\% \\
Embed., max-pair & $\checkmark$ & \textbf{9.6\%} & \textbf{0.4\%} \\
\bottomrule
\end{tabular}
\end{table}

\subsection{Genericization Stress Test}
\label{app:genericization}

A policy could lower checkable visual inconsistencies by collapsing to source-free, generic appearance edits that repeat across images. We test whether the reward composition permits this shortcut.

The stress policy optimizes the click RM as its sole learned signal. It drops the PPL and diversity rewards retained by the preference row of Table~\ref{tab:rewardablation}.

Relative to the SFT reference, the mean click-RM score rises by $+0.31$, $+1.25$, and $+2.46$ after 30, 80, and 160 steps. Meanwhile, the policy collapses. At step 160, cross-image reuse reaches 99.9\%, whereas within-list redundancy remains 0.1\%: essentially one generic edit is reused across nearly every image. An unconstrained per-suggestion attractiveness model therefore rewards this degenerate solution.

\begin{table}[t]
\caption{RM-only genericization stress test. Cross-image reuse and within-list redundancy are measured at the final checkpoint.}
\label{tab:genericization}
\reportfloatwidth{0.50}
\small
\setlength{\tabcolsep}{6pt}
\begin{tabular}{@{}rccc@{}}
\toprule
Steps & $\Delta$ click-RM & Reuse $\downarrow$ & Redund. $\downarrow$\\
\midrule
30  & $+0.31$ & -- & --\\
80  & $+1.25$ & -- & --\\
160 & $+2.46$ & 99.9\% & 0.1\%\\
\bottomrule
\end{tabular}
\end{table}

This failure motivates the PPL reward, max-pair diversity, and cross-image reuse audit. Under the full reward set, the full framework has 54.3\% reuse versus 57.3\% for SFT, with similar length variation (2.9 versus 3.0 characters). The deployed policy therefore does not take the generic route.

\section{Human Evaluation Protocol}
\label{app:human}

Eight experts evaluate randomized policy outputs while seeing the latest image, current query, and current intent, but no reward scores. Every arm is evaluated on the same 800 suggestion judgments.

Following the production annotation worksheet, every suggestion receives one integer quality score:
\begin{itemize}[leftmargin=*,label={},topsep=2pt,itemsep=1pt]
\item \textbf{3 (excellent)}: a natural continuation of the user's intent that refers to real image content, is directly executable, and has clear editing value;
\item \textbf{2 (usable)}: relevant and executable, but less valuable, concrete, or natural as a continuation;
\item \textbf{1 (unusable)}: weakly connected to the intent, an abrupt turn, redundant with the request or current image, or otherwise inappropriate for the scene;
\item \textbf{0 (severe failure)}: conflicts with an explicit user constraint, relies on nonexistent image content, is logically malformed, or violates a product rule.
\end{itemize}

Following the score-difference aggregation convention of prior deployed query-suggestion evaluation \cite{cici2025clicks}, but using our suggestion-level 0--3 rubric, GSB is the difference between the summed expert scores of a policy and PE under the common blind evaluation protocol. PE receives the same raw ratings as every other arm; its GSB is zero because it is the reporting reference. With 800 judgments, the range is $[-2400,2400]$, and a positive value is the net expert score gained over the launch policy. Unscored or absent worksheet entries do not enter any arm's total.

The worksheet also records a 0--2 within-slate diversity score, structured issue tags, free-text rationales, annotator identity, and a separate quality-control decision. A second reviewer checks completed annotations and resolves requested corrections before aggregation.

A score of 0 may combine visual hallucination with other severe failures. Verifier quality is therefore measured on the calibration benchmark in Section~\ref{subsec:verifierstudy}, not inferred from GSB.

\section{Qualitative Analysis}
\label{app:qualitative}

Held-out cases illustrate the quantitative failure modes. For a desk figurine, SFT incorrectly restyles the object as a person, whereas the full framework proposes changes to the office lighting and layout. For a document image, the full framework refers to the visible red seal instead of suggesting an unrelated background edit. After a request for a slimmer body, it avoids a contradictory broader-shoulder suggestion. On a mathematics grid, it proposes related area and perimeter exercises rather than a generic enhancement. These cases test visible-object use, directional consistency, and usefulness rather than fluency alone.

Three recurring failures remain. The model can emit polished but low-value paraphrases, miss subtle visual states when the verifier fails open, and reject rare but valid intent transitions. Max-pair diversity and the PPL reward partly address paraphrase collapse, and validated general backfills reduce empty slates when a rare transition is blocked.

\section{Online Experiment Details}
\label{app:online}

The online experiment runs for 14 days on the Qwen App image-generation surface. All arms run concurrently on the same eligible population. Assignment is randomized and fixed at the user level, so one user sees the same policy across sessions. Each policy arm receives the same 5\% traffic allocation and includes millions of users. In every arm, the policy generates a candidate slate and the shared display layer randomly selects three suggestions without replacement. PE runs concurrently as the common control arm for SFT, Stage 2, and the full framework; the recommendation policy is the only experimental difference.

Recommendation CTR on the three-item display slate, image take-away rate, and average conversation turns per user are the three core online metrics. Latency, generation failure, and negative feedback are monitored as operational constraints, and a regression on any of them blocks a launch regardless of engagement lifts. Table~\ref{tab:main} reports PE-relative lifts; all are significant ($p<0.05$).

\end{document}